\documentclass[journal]{IEEEtran}
\usepackage{eurosym}
\usepackage{cite,algorithm,array,url}
\usepackage{graphicx,colortbl}
\usepackage{amsfonts,times,amsmath,amssymb,amsthm}
\usepackage{multirow,booktabs}
\usepackage{color}
\usepackage{algorithm}
\usepackage{algpseudocode}
\usepackage{tikz}
\usepackage{lscape}
\usepackage{multirow}

\ifCLASSOPTIONcompsoc
\usepackage[caption=false,font=normalsize,labelfon
t=sf,textfont=sf]{subfig}
\else
\usepackage[caption=false,font=footnotesize]{subfi
g}
\fi

\newcolumntype{L}[1]{>{\raggedright\arraybackslash}p{#1}}
\newcolumntype{C}[1]{>{\centering\arraybackslash}m{#1}}
\newcolumntype{R}[1]{>{\raggedleft\arraybackslash}m{#1}}
\newcolumntype{P}[1]{>{\centering\arraybackslash}p{#1}}

\graphicspath{{images/}}

\theoremstyle{plain}

\theoremstyle{definition}

\theoremstyle{remark}

\begin{document}

\title{A Domain-Knowledge-Aided Deep Reinforcement Learning Approach for Flight Control Design
}
\author{Hyo-Sang~Shin, Shaoming~He and Antonios~Tsourdos \thanks{
Hyo-Sang Shin and Antonios Tsourdos are with the School of Aerospace, Transport and Manufacturing, Cranfield University, Cranfield MK43 0AL, UK
(email: \texttt{\{h.shin,a.tsourdos\}@cranfield.ac.uk})}
\thanks{
Shaoming He is with the School of Aerospace Engineering, Beijing Institute of Technology, Beijing 100081, China
(email: \texttt{shaoming.he@bit.edu.cn}) and also with the School of Aerospace, Transport and Manufacturing, Cranfield University, Cranfield MK43 0AL, UK
(email: \texttt{shaoming.he@cranfield.ac.uk})}
\thanks{``This work has been submitted to the IEEE for possible publication. Copyright may be transferred without notice, after which this version may no longer be accessible.''}
}

\maketitle

\begin{abstract}
This paper aims to examine the potential of using the emerging deep reinforcement learning techniques in flight control. Instead of learning from scratch, we suggest to leverage domain knowledge available in learning to improve learning efficiency and generalisability. More specifically, the proposed approach fixes the autopilot structure as typical three-loop autopilot and deep reinforcement learning is utilised to learn the autopilot gains. To solve the flight control problem, we then formulate a Markovian decision process with a proper reward function that enable the application of reinforcement learning theory. Another type of domain knowledge is exploited for defining the reward function, by shaping reference inputs in consideration of important control objectives and using the shaped reference inputs in the reward function. The state-of-the-art deep deterministic policy gradient algorithm is utilised to learn an action policy that maps the observed states to the autopilot gains. Extensive empirical numerical simulations are performed to validate the proposed computational control algorithm.
\end{abstract}


\section{Introduction}

The main objective of a flight controller for modern air vehicles is to track a given command in a fast and stable manner. Classical linear autopilot in conjunction with gain scheduling technique is one of widely-accepted frameworks for flight controller design due to its simplicity, local stability and ease of implementation \cite{zarchan2012tactical,stilwell2001state,stilwell1999interpolation,theodoulis2009missile,lhachemi2016gain}. This technique requires to linearise the airframe dynamics around several characteristic trim points and a static feedback linear controller is designed for each operation point. The controller gains are then online scheduled through a proper interpolation algorithm to cover the entire flight envelop. 

The systematic gain-scheduling approach provides engineers an intuitive framework to design simple and effective controllers for nonlinear airframes. The issue is that its performance might be significantly degraded for a highly non-linear and coupled system in which the assumptions on the conventional linear control theory could be violated. To resolve the issue, there have been extensive studies on other control theories, e.g., sliding mode control \cite{thukral1998sliding,shkolnikov2000robust}, backstepping \cite{mattei2014nonlinear}, adaptive control \cite{calise2000adaptive,wang2008l1}, state-dependent Riccati equation (SDRE) method \cite{mracek1997full,mracek2007sdre} and $H_\infty$ synthesis \cite{buschek2003design,kim2017augmented}.  However, each control method has its own advantages and limitations. For example, sliding mode control usually suffers from the chattering problem and therefore it is difficult to implement in practice. The backstepping autopilot requires to calculate the derivatives of the virtual commands, which normally contain some information that cannot be directly measured. The implementation of SDRE controller requires to solve the complicated algebraic Riccati equation at each sampling instant. In a recent contribution \cite{lee2016connections}, nonlinear flight controllers have been proved to share the same structure with linear gain-scheduling controllers by properly adjusting the feedback gains and therefore might suffer from similar drawbacks: requiring partial model information in controller design.

Thanks to the rapid development on embedded computational capability, there has been an increasing attention on the development of computational control or numerical optimisation algorithms in recent years \cite{lu2017introducing}. Unlike classical optimal autopilot, computational control algorithms generate the control input relies extensively on onboard computation and there is no analytic solution of any specific control law. Generally, computational control algorithms can be classified into two main categories: (1) model-based ; and (2) data-based. The authors in \cite{tang2012predictive,bachtiar2014nonlinear,bachtiar2017nonlinear} leveraged model predictive control (MPC) to design a robust autopilot for agile airframes. The basic idea behind MPC is that it solves a constrained nonlinear optimisation problem at each time instant in a receding horizon manner and therefore shows appealing advantages in autopilot design. Except for MPC, bio-inspired numerical optimisation algorithms, e.g., genetic algorithm, particle swarm optimisation, have also been reported for flight controller design in recent years \cite{krishnakumar1992control,karimi2011multivariable}.

Most of the flight control algorithms discussed so far are model-based control algorithms: they are generally designed under the assumption that the model information is correctly known. It is clear that the performance of model-based optimisation approaches highly relies on the accuracy of the model utilised. For modern air vehicles that suffer from aerodynamic uncertainties, it would be more beneficial to develop data-based autopilot. Considering the properties of the autopilot problem, leveraging the reinforcement learning (RL) concept might be an attractive option for developing a data-based control algorithm \cite{wang2017adaptive,li2019transforming,tan2019cooperative,Koch_2019,huang2017parameterized,ding2018adaptive,cui2017adaptive}. 

To this end, this paper aims to examine the potential of using the emerging deep RL (DRL) techniques in flight controller design. The problem is formulated in the RL framework by defining a proper Markovian Decision Process (MDP) along with a reward function. Since the problem considered is a continuous-time control problem, the state-of-the-art policy gradient algorithm, i.e., deep deterministic policy gradient (DDPG), is utilised to learn a deterministic action function.

Previous works using RL or DRL to solve flight control problems mainly try to obtain networks, mapping from states directly to control actions \cite{ferrari2004online,enns2003helicopter,zhou2018incremental,wang2019deterministic,xu2019morphing,wu2018depth}. As initial investigations, we followed previous relevant studies, applied them in an aerial vehicle model for low-level attitude control and examined their performance. From the investigation, it is found that the learning effectiveness is generally an issue and it is extremely difficult to generalise, i.e. to train networks in a way providing a reasonable performance to different types of inputs to the control system. Our initial investigation suggests that this becomes more problematic in the aerial vehicle whose dynamics is statically unstable as a small change in the control command could result in an unstable response.

An extensive literature survey indicated that there have some efforts to utilise prior or domain knowledge to improve learning effectiveness, generalisation or both \cite{shapiro2001using,Chen-2019-112809,gulccehre2016knowledge,ferranti2017value,moreno2004using}. In \cite{shapiro2001using}, the authors proposed an learning architecture by utilising background knowledge, i.e., an approximate model of the behaviour, in reinforcement learning and demonstrated that this concept enables fast learning with reduced size of learning samples. The results of \cite{Chen-2019-112809} revealed that using prior knowledge of the tasks can significantly boost the learning performance and generalisation capabilities. In \cite{gulccehre2016knowledge,ferranti2017value,moreno2004using}, the authors suggested a particular structure that guides the learner by incorporating prior knowledge in supervised learning and reinforcement learning. This concept is proved to help increase the learning efficiency and avoid getting stuck of the training process both in terms of training and generalisation error. The results from these studies strongly suggest that leveraging domain knowledge can improve the learning effectiveness. 

To this end, this paper focuses on developing a DDPG-based control framework that exploits the domain knowledge to enhance the learning efficiency. The main contributions of the development are three aspects: 

(1) We develop a DDPG-based control framework that utilises the domain knowledge, that is the control structure. As discussed, most of RL or Deep RL (DRL) based algorithms directly learn control actions from scratch that might hinder the learning efficiency. Unlike these typical algorithms, the proposed framework is formulated by fixing the autopilot structure. Under the problem formulated, DRL learns a deterministic action function that maps the observed engagements states to the autopilot gains. This enables the DRL-based control to enhance the learning efficiency, retain the strengths of simple structure and improve the performance of classical autopilot via learning. 

(2) The reference input to the control system is shaped by considering several important criteria, e.g., rising time, damping ratio, overshooting, in control system design. Then, the shaped reference input is leveraged in the reward function. This greatly simplifies the parameter tuning process in Multi-Objective Optimisation (MOO). Note that there are many objectives required to be achieved in control system design and hence control problems are often expressed as an MOO problem. The proposed approach also allows the DRL-based control to resolve the particularity in applying RL/DRL to the control system design problem.  

(3) In the proposed DDPG-based control framework, this paper suggests to use normalised observations and actions in the training process to tackle the issue with the scale of rewards and networks. Our initial investigation indicated that different scales in states create different scales in rewards and networks. Consequently, the training becomes ineffective, which will be shown via numerical simulations in Section \ref{sub:test_result}. The numerical simulations confirm that this issue can be resolved by the normalisation.

The proposed concepts and performance are examined through extensive numerical studies. The numerical analysis reveals that the proposed DDPG autopilot guarantees satisfactory tracking performance and exhibits several advantages against traditional gain scheduling approach. Robustness of the domain-knowledge-aided DDPG autopilot is also investigated by examining the performance of the proposed approach against model uncertainty. The simulation results on the robustness examination confirm that the DDPG autopilot developed is robust against model uncertainty. Also, relative stability of the proposed autopilot is numerically investigated and its results show that the proposed autopilot meets the typical design criteria on the phase and gain margins. 

The rest of the paper is organised as follows. Sec. II introduces of the basic concept of deep reinforcement learning. Sec. III presents nonlinear dynamics of airframes, followed by the details of the proposed computational flight control algorithm in Sec. IV. Finally, some numerical simulations and conclusions are offered. 

\section{Deep Reinforcement Learning}

For the completeness of this paper, this section presents some backgrounds and preliminaries of reinforcement learning and DDPG.

\subsection{Reinforcement Learning}

In the RL framework, an agent learns an action policy through episodic interaction with an unknown environment. The RL problem is often formalised as a MDP or a partially observable MDP (POMDP). A MDP is described by a five-tuple $\left(\mathcal S, \mathcal O, \mathcal A, \mathcal P, \mathcal R\right)$, where $\mathcal S$ refers to the set of states,  $\mathcal O$ the set of observations, $\mathcal A$ the set of actions, $\mathcal P$ the state transition probability and $\mathcal R$ the reward function. If the process is fully observable, we have $\mathcal S = \mathcal O$. Otherwise, $\mathcal S \ne \mathcal O$. 

At each time step $t$, an observation $o_t \in \mathcal O$ is generated from the internal state $s_t \in \mathcal S$ and given to the agent. The agent utilises this observation to generate an action $a_t \in \mathcal A$ that is sent to the environment, based on specific action policy $\pi$. The action policy is a function that maps observations to a probability distribution over the actions.  The environment then leverages the action and the current state to generate the next state $s_{t+1}$ with conditional probability $\mathcal P \left( s_{t+1} | s_t,a_t \right)$ and a scalar reward signal $r_t \sim \mathcal R \left( s_t,a_t \right) $. For any trajectory in the state-action space, the state transition in RL is assumed to follow a stationary transition dynamics distribution with conditional probability satisfying the Markov property, i.e.,
\begin{equation} \label{eq:1} 
\mathcal P \left( s_{t+1} | s_1,a_1, \cdots, s_t,a_t \right)=\mathcal P \left( s_{t+1} | s_t,a_t \right)
\end{equation} 
The goal of RL is to seek a policy for an agent to interact with an unknown environment while maximising the expected total reward it received over a sequence of time steps. The total reward in RL is defined as the summation of discounted reward as
\begin{equation} \label{eq:2} 
R_t=\sum\limits_{i = t}^N {{\gamma ^{i - t}}{r_i}} 
\end{equation} 
where $\gamma \in \left( 0,1 \right]$ denotes the discounting factor. 

Given current state $s_t$ and action $a_t$, the expected total reward is also known as the action-value function
\begin{equation} 
Q^{\pi} \left( s_t,a_t \right)= \mathbb E _{\pi}\left[ R_t| s_t,a_t\right]
\end{equation}
which satisfies a recursive form as
\begin{equation} \label{eq:5}
Q^{\pi} \left( s_t,a_t \right)=\mathbb E _{\pi}\left[ \mathcal R \left( s_t,a_t \right)  +  \gamma \mathbb E _{\pi}\left[ Q^{\pi} \left( s_{t+1},a_{t+1} \right)\right]  \right]
\end{equation}
Therefore, the optimal policy can be obtained by optimising the action-value function. However, directly optimising the action-value function or action-value function requires accurate model information and therefore is difficult to implement with model uncertainties. Model-free RL algorithms relax the requirement on accurate model information and hence can be utilised even with high model uncertainties.


\subsection{Deep Deterministic Policy Gradient}

\begin{figure}
\begin{center}

\tikzset{every picture/.style={line width=0.7pt}} 

\begin{tikzpicture}[x=0.75pt,y=0.75pt,yscale=-1,xscale=1, scale=0.65]

\draw  [color={rgb, 255:red, 245; green, 166; blue, 35 }  ,draw opacity=1 ][line width=2.25]  (100,182.4) .. controls (100,176.1) and (105.1,171) .. (111.4,171) -- (206.1,171) .. controls (212.4,171) and (217.5,176.1) .. (217.5,182.4) -- (217.5,216.6) .. controls (217.5,222.9) and (212.4,228) .. (206.1,228) -- (111.4,228) .. controls (105.1,228) and (100,222.9) .. (100,216.6) -- cycle ;
\draw  [color={rgb, 255:red, 208; green, 2; blue, 27 }  ,draw opacity=1 ][line width=2.25]  (300,76.4) .. controls (300,70.1) and (305.1,65) .. (311.4,65) -- (406.1,65) .. controls (412.4,65) and (417.5,70.1) .. (417.5,76.4) -- (417.5,110.6) .. controls (417.5,116.9) and (412.4,122) .. (406.1,122) -- (311.4,122) .. controls (305.1,122) and (300,116.9) .. (300,110.6) -- cycle ;
\draw  [color={rgb, 255:red, 189; green, 16; blue, 224 }  ,draw opacity=1 ][line width=2.25]  (301,182.4) .. controls (301,176.1) and (306.1,171) .. (312.4,171) -- (407.1,171) .. controls (413.4,171) and (418.5,176.1) .. (418.5,182.4) -- (418.5,216.6) .. controls (418.5,222.9) and (413.4,228) .. (407.1,228) -- (312.4,228) .. controls (306.1,228) and (301,222.9) .. (301,216.6) -- cycle ;
\draw [color={rgb, 255:red, 208; green, 2; blue, 27 }  ,draw opacity=1 ][line width=2.25]    (358.5,170.38) -- (359.42,125) ;
\draw [shift={(359.5,121)}, rotate = 451.16] [color={rgb, 255:red, 208; green, 2; blue, 27 }  ,draw opacity=1 ][line width=2.25]    (17.49,-5.26) .. controls (11.12,-2.23) and (5.29,-0.48) .. (0,0) .. controls (5.29,0.48) and (11.12,2.23) .. (17.49,5.26)   ;

\draw [color={rgb, 255:red, 189; green, 16; blue, 224 }  ,draw opacity=1 ][line width=2.25]    (218.5,198) -- (298.5,198) ;
\draw [shift={(302.5,198)}, rotate = 180] [color={rgb, 255:red, 189; green, 16; blue, 224 }  ,draw opacity=1 ][line width=2.25]    (17.49,-5.26) .. controls (11.12,-2.23) and (5.29,-0.48) .. (0,0) .. controls (5.29,0.48) and (11.12,2.23) .. (17.49,5.26)   ;

\draw [color={rgb, 255:red, 189; green, 16; blue, 224 }  ,draw opacity=1 ][line width=2.25]    (417.5,198) -- (497.5,198) ;
\draw [shift={(501.5,198)}, rotate = 180] [color={rgb, 255:red, 189; green, 16; blue, 224 }  ,draw opacity=1 ][line width=2.25]    (17.49,-5.26) .. controls (11.12,-2.23) and (5.29,-0.48) .. (0,0) .. controls (5.29,0.48) and (11.12,2.23) .. (17.49,5.26)   ;

\draw [color={rgb, 255:red, 245; green, 166; blue, 35 }  ,draw opacity=1 ][line width=2.25]    (451,199) -- (450.5,263) ;

\draw [color={rgb, 255:red, 245; green, 166; blue, 35 }  ,draw opacity=1 ][line width=2.25]    (48.05,262.74) -- (450.5,263) ;

\draw [color={rgb, 255:red, 245; green, 166; blue, 35 }  ,draw opacity=1 ][line width=2.25]    (48.56,198.74) -- (48.05,262.74) ;

\draw [color={rgb, 255:red, 245; green, 166; blue, 35 }  ,draw opacity=1 ][line width=2.25]    (48.56,198.74) -- (96.5,198.98) ;
\draw [shift={(100.5,199)}, rotate = 180.29] [color={rgb, 255:red, 245; green, 166; blue, 35 }  ,draw opacity=1 ][line width=2.25]    (17.49,-5.26) .. controls (11.12,-2.23) and (5.29,-0.48) .. (0,0) .. controls (5.29,0.48) and (11.12,2.23) .. (17.49,5.26)   ;

\draw [color={rgb, 255:red, 245; green, 166; blue, 35 }  ,draw opacity=1 ][line width=2.25]    (160.5,96) -- (299.5,95) ;

\draw [color={rgb, 255:red, 245; green, 166; blue, 35 }  ,draw opacity=1 ][line width=2.25]    (160.5,96) -- (160.5,168) ;
\draw [shift={(160.5,172)}, rotate = 270] [color={rgb, 255:red, 245; green, 166; blue, 35 }  ,draw opacity=1 ][line width=2.25]    (17.49,-5.26) .. controls (11.12,-2.23) and (5.29,-0.48) .. (0,0) .. controls (5.29,0.48) and (11.12,2.23) .. (17.49,5.26)   ;

\draw (158.75,199.5) node [rotate=-0.8] [align=left] {\textbf{\small{Actor (Provide}}\\\textbf{\small{Control Policy)}}};
\draw (358.75,93.5) node [rotate=-0.8] [align=left] {\textbf{\footnotesize{Critic (Evaluate}}\\\textbf{\footnotesize{Control Policy)}}};
\draw (359.75,198.5) node [rotate=-0.8] [align=left] {\textbf{Environment}};
\draw (398,146) node  [align=left] {\textbf{Reward}};
\draw (459,182) node  [align=left] {\textbf{Output}};
\draw (272,248) node  [align=left] {\textbf{Observation}};
\draw (195,83) node  [align=left] {\textbf{Performance Evaluation}};
\draw (262,184) node  [align=left] {\textbf{Action}};

\end{tikzpicture}

\end{center}
\caption{Basic concept of DDPG.}
\label{fig:ddpg}
\end{figure}
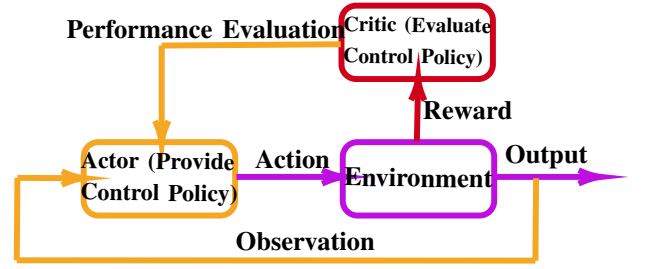

For the autopilot problem, the main goal is to find a deterministic actuator command that could drive the air vehicle to track the target lateral acceleration command in a rapid and stable manner. Since this is a continuous control problem, we utilise the the model-free policy gradient RL approach to learn a deterministic function that directly maps the states to the actions, i.e., the action function is updated by following the gradient direction of the value function with respect to the action, thus termed as policy gradient. More specifically, the state-of-the-art policy-gradient solution, DDPG proposed by Google Deepmind \cite{lillicrap2015continuous}, is leveraged to develop a computational lateral acceleration autopilot for an air vehicle. DDPG is an Actor-Critic method that consists of two main function blocks: (1) Critic evaluates the given policy based on current states to calculate the action-value function; (2) Actor generates policy based on the evaluation of critic. DDPG utilises two different deep neural networks, i.e., actor network and critic network, to approximate the action function and the action-value function. The basic concept of DDPG is shown in Fig. \ref{fig:ddpg}.

Denote $A^\mu \left(s_t\right)$ as the deterministic policy, which is a function that directly maps the states to the actions, i.e., $a_t = A^\mu \left(s_t\right)$. Here, we assume that the action function $A^\mu \left(s_t\right)$ parameterised by $\mu$. In DDPG, the actor function is optimised by adjusting the parameter $\mu$ toward the gradient of the expected total reward as \cite{lillicrap2015continuous}
\begin{equation} \label{eq:6}
{\nabla _{{a_t}}}{Q^w}\left( s_t,A^\mu\left( s_t \right) \right)={\nabla _\mu }{A^\mu }\left( {{s_t}} \right){\nabla _{{a_t}}}{Q^w}\left( {{s_t},{a_t}} \right)
\end{equation}
where $Q^w \left( s_t,a_t \right)$ stands for the action-value function, which is parameterised by $w$.

The parameter $\mu$ is then updated by moving the policy in the direction of the gradient of $Q^w$ in a recursive way as 
\begin{equation} \label{eq:7}
\mu_{t+1} = \mu_t +\alpha_\mu {\nabla _\mu }{A^\mu }\left( {{s_t}} \right){\nabla _{{a_t}}}{Q^w}\left( {{s_t},{a_t}} \right)
\end{equation}
where $\alpha_\mu$ refers to the learning rate of the actor network.

Similar to Q-learning, DDPG also utilises the temporal-difference (TD) error $\delta_t$ in approximating the error of action-value function as
\begin{equation} \label{eq:8}
\delta_t = r_t + \gamma {Q^w}\left( s_{t+1},A^\mu \left(s_{t+1}\right) \right) - {Q^w}\left( {{s_t},{a_t}} \right)
\end{equation}

DDPG utilises the square of TD error as the loss function $\mathcal L (w)$ in updating the critic network, i.e.,
\begin{equation}
\mathcal L (w) = \delta_i^2
\end{equation}

Taking the partial derivative of $\mathcal L (w)$ respect to $w$ gives
\begin{equation}
\nabla _w \mathcal L (w)= - 2\delta_i  \nabla _w {Q^w}\left( {{s_t},{a_t}} \right)
\end{equation}

The parameter $w$ is then updated using gradient descent by following the negative gradient of  $\mathcal L (w)$ as
\begin{equation} \label{eq:9}
w_{t+1} = w_t +\alpha_w \delta_t \nabla _w {Q^w}\left( {{s_t},{a_t}} \right)
\end{equation}
where $\alpha_w$ stands for the learning rate of the critic network.

One major issue of using deep neural networks in RL is that most neural network optimisation algorithms assume that the samples for training are independently and identically distributed. However, this assumption is violated if the training samples are directly generated by sequentially exploring the environment. To resolve this issue, DDPG leverages a mini batch buffer that stores the training samples using the experience replay technique. Denote $e_t = \left( s_t, a_t, r_t, s_{t+1}\right)$ as the transition experience of the $t$th step. DDPG utilises a buffer $\mathcal D$ with its size being $|\mathcal D|$ to store transition experiences. DDPG stores the current transition experience in the buffer and deletes the oldest one if the number of the transition experience reaches the maximum value $|\mathcal D|$. At each time step during training, DDPG uniformly draws $N$ transition experience samples from the buffer $\mathcal D$ and utilises these random samples to train actor and critic networks. By utilising the experience buffer, the critic network is updated as
\begin{equation} \label{eq:12} 
\begin{split}
{\nabla _\mu }J\left( {{A^\mu }} \right) &= \frac{1}{N}\sum\limits_{i = 1}^N {{\nabla _\mu }{A^\mu }\left( {{s_i}} \right){\nabla _{{a_t}}}{Q^w}\left( {{s_i},{a_i}} \right)}  \\
\mu_{t+1} &= \mu_t +\alpha_\mu {\nabla _\mu }J\left( {{A^\mu }} \right)
\end{split}
\end{equation}

With $N$ transition experience samples, the loss function in updating the critic network now becomes
\begin{equation} \label{eq:13}
\mathcal L (w) =  \frac{1}{N}\sum\limits_{i = 1}^N {\delta_i^2}
\end{equation}

The parameter of the critic network is updated by gradient decent as
\begin{equation} \label{eq:14} 
\begin{split}
{\nabla _w }\mathcal L\left( w \right) &=\frac{1}{N}\sum\limits_{i = 1}^N {\delta_i \nabla _w {Q^w}\left( {{s_i},{a_i}} \right)}  \\
w_{t+1} &= w_t +\alpha_w {\nabla _w }\mathcal L(w)
\end{split}
\end{equation}

Notice that the update of the action-value function is also utilised as the target value as shown in Eq. (\ref{eq:8}), which might cause the divergence of critic network training \cite{lillicrap2015continuous}. To address this problem, DDPG creates one target actor network and one target critic network. Suppose the additional actor and critic networks are paramterised by $\mu'$ and $w'$, respectively. These two target networks use soft update, rather than directly copying the parameters from the original actor and critic networks, as
\begin{equation} \label{eq:10} 
\begin{split}
\mu' &= \tau \mu + \left(1-\tau\right) \mu' \\
w' &= \tau w + \left(1-\tau\right) w' 
\end{split}
\end{equation}
where $\tau \ll  1$ is a small updating rate. This soft update law shares similar concept as low-frequency learning in model reference adaptive control to improve the robustness of the adaptive process \cite{yucelen2012low,gaudio2019connections}.

The soft-updated two target networks are then utilised in calculating the TD-error as
\begin{equation} \label{eq:11}
\delta_t = r_t + \gamma {Q^{w'}}\left( s_{t+1},A^{\mu'} \left(s_{t+1}\right) \right) - {Q^w}\left( {{s_t},{a_t}} \right)
\end{equation}

With very small update rate, the stability of critic network training greatly improves at the expense of slow training process. Therefore, the update rate is a tradeoff between training stability and convergence speed.


The detailed pseudo code of DDPG is summarised in Algorithm \ref{algo:ddpg}.

\begin{algorithm}
\caption{Deep deterministic policy gradient}\label{algo:ddpg}
\begin{algorithmic}[1]
\State Initialise the actor and critic networks with random weights $\mu$ and $w$
\State Initialise the target actor and critic networks with weights $\mu' \leftarrow \mu$ and $w' \leftarrow w$
\State Initialise the experience buffer $\mathcal D$
\For {episode = 1: MaxEpisode}
	\For {$t= 1: \text{MaxStep}$}
		\State Generate an action from the actor network based on current state $a_t = A^\mu(s_t)$
		\State Add a random noise $v_t$ to the action for exploration $a_t' = a_t +v_t$
		\State Execute the action $a_t'$ and receive new state $s_{t+1}$ and reward $r_t$ 
		\State Store the transition experience $e_t = \left( s_t, a_t, r_t, s_{t+1}\right)$ in the experience buffer $\mathcal D$
		\State Uniformly draw $N$ random samples $e_i$ from the experience buffer $\mathcal D$
		\State Calculate the TD error $\delta_i$
\[\delta_i = r_i+ \gamma {Q^{w'}}\left( s_{i+1},A^{\mu'} \left(s_{i+1}\right) \right) - {Q^w}\left( {{s_i},{a_i}} \right)\]
		\State Calculate the loss function $\mathcal L (w)$
\[\mathcal L (w) =  \frac{1}{N}\sum\limits_{i = 1}^N {\delta_i^2}\]
		\State Update the critic network using gradient descent as
\[\begin{split}
{\nabla _w }\mathcal L\left( w \right) &=\frac{1}{N}\sum\limits_{i = 1}^N {\delta_i \nabla _w {Q^w}\left( {{s_i},{a_i}} \right)}  \\
w_{t+1} &= w_t +\alpha_w {\nabla _w }\mathcal L(w)
\end{split}\]
		\State Update the actor network using policy gradient as
\[\begin{split}{\nabla _\mu }J\left( {{A^\mu }} \right) &= \frac{1}{N}\sum\limits_{i = 1}^N {{\nabla _\mu }{A^\mu }\left( {{s_i}} \right){\nabla _{{a_t}}}{Q^w}\left( {{s_i},{a_i}} \right)}  \\
\mu_{t+1} &= \mu_t +\alpha_\mu {\nabla _\mu }J\left( {{A^\mu }} \right)
\end{split}\]
		\State Update the target networks as
\[\begin{split}
\mu' &= \tau \mu + \left(1-\tau\right) \mu' \\
w' &= \tau w + \left(1-\tau\right) w' 
\end{split}\]
		\If {the task is accomplished}
			\State Terminate the current episode
                      \EndIf
	\EndFor
\EndFor
\end{algorithmic}
\end{algorithm}

\section{Nonlinear Airframe Dynamics Model}

\begin{figure}[tb]
\centering
\resizebox{80mm}{!}{\includegraphics{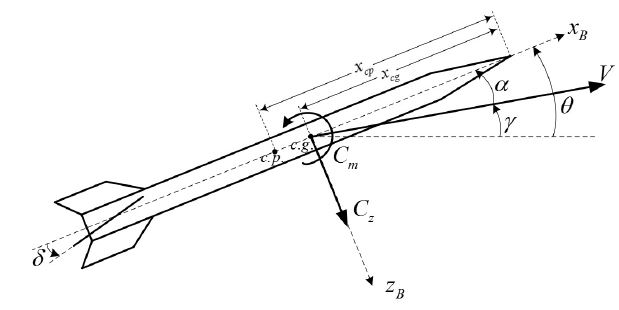}}
\caption{The longitudinal dynamics model and parameter definitions.}
\label{fig:dynamics}
\end{figure}

For the feasibility investigation of the proposed approach, this paper utilises the longitudinal dynamics model of a tail-controlled skid-to-turn airframe in autopilot design \cite{mracek1997full}, as shown in Fig. \ref{fig:dynamics}. For simplicity, we assume that the air vehicle is roll-stabilised, e.g., zero roll angle and zero roll rate, and has constant mass, i.e., after boost phase. Under these assumptions, the nonlinear longitudinal dynamics model can be expressed as
\begin{equation}
\begin{split}
 &\dot \alpha  = \frac{{QS}}{{mV}}\left( {{C_N}\cos \alpha  - {C_A}\sin \alpha } \right) + \frac{g}{V}\cos \gamma  + q \\ 
& \dot q = \frac{{QSd}}{{{I_{yy}}}}{C_M} \\ 
& \dot \theta = q \\
& \gamma = \theta - \alpha \\
& a_z = V \dot \gamma
 \end{split}
\label{eq:15}
\end{equation}
where the parameters $\alpha$, $\theta$, $\gamma$ and $a_z$ represent angle-of-attack, pitch attitude angle, flight path angle and lateral acceleration, respectively.  In Eq. \eqref{eq:15}, $m$, $g$ and $V$ stand for mass, gravitational acceleration and velocity, respectively. The variable $Q$ represents the dynamic pressure, which is defined as $Q=0.5\rho V^2$ with $\rho$ being the air density. Additionally, the parameters $S$, $d$, $I_{yy}$, and $m$ denote reference area, diameter, moment of inertia, and mass, respectively. The values of all physical parameters are detailed in Table \ref{tab:1}.

\begin{table}[tb]
\begin{center}
\caption{Physical parameters.}
\label{tab:1}
\begin{tabular}{ccc}
\hline\hline
Symbol & Name & Value \\ \hline
$I_{yy}$ & Moment of Inertia & 247.439 $kg\cdot m^2$\\
$S$ &Reference Area & 0.0409 $m^2$\\
$d$ & Reference Distance & 0.2286 $m$\\
$m$ & Mass & 204.02 $kg$\\
$g$ & Gravitational acceleration & 9.8 $m/s^2$\\
\hline\hline
\end{tabular}
\end{center}
\end{table}

The aerodynamic coefficients $C_A$, $C_N$ and $C_M$ are determined as
\begin{equation}
\begin{split}
& {C_A} = {a_a} \\ 
& {C_N} = {a_n}{\alpha ^3} + {b_n}\alpha \left| \alpha  \right| + {c_n}\left( {2 - \frac{M}{3}} \right)\alpha  + {d_n}\delta  \\ 
& {C_M} = {a_m}{\alpha ^3} + {b_m}\alpha \left| \alpha  \right| + {c_m}\left( { - 7 + \frac{{8M}}{3}} \right)\alpha  + {d_m}\delta  \\ 
 \end{split}
\label{eq:16}
\end{equation}
where $a_i$, $b_i$, $c_i$ and $d_i$ with $i = a,n,m$ are constants and the values are presented in Table \ref{tab:2}. The parameters $M$ and $\delta$ represent Mach number and control fin deflection, respectively. The Mach number is subject to the following differential equation
\begin{equation}
\dot M = \frac{{QS}}{{mV_s}}\left( {{C_N}\sin \alpha  + {C_A}\cos \alpha } \right) - \frac{g}{V_s}\sin \gamma
\label{eq:17}
\end{equation}
where $V_s$ is the speed of sound.

\begin{table}[tb]
\begin{center}
\caption{Aerodynamic polynomial coefficients.}
\label{tab:2}
\begin{tabular}{cccc}
\hline\hline
Symbol & Value & Symbol & Value \\ \hline
$a_a$ & 0.3 & $a_m$ & 40.44\\
$a_n$ & 19.373 & $b_m$ & -64.015\\
$b_n$ & -31.023 & $c_m$ & 2.922\\
$c_n$ & -9.717 & $d_m$ & -11.803\\
$d_n$ &-1.948\\
\hline\hline
\end{tabular}
\end{center}
\end{table}

The actuator of an air vehicle is usually modelled by a second-order dynamic system as 
\begin{equation}
\left[ {\begin{array}{*{20}{c}}
   {\dot \delta }  \\
   {\ddot \delta }  \\
\end{array}} \right] = \left[ {\begin{array}{*{20}{c}}
   0 & 1  \\
   { - \omega _a^2} & { - 2{\xi _a}{\omega _a}}  \\
\end{array}} \right]\left[ {\begin{array}{*{20}{c}}
   \delta   \\
   {\dot \delta }  \\
\end{array}} \right] + \left[ {\begin{array}{*{20}{c}}
   0  \\
   {\omega _a^2}  \\
\end{array}} \right]{\delta _c}
\label{eq:18}
\end{equation}
where ${\xi _a}=0.7$ and ${\omega _a}=150 rad/s$ denote the damping ratio and natural frequency, respectively. The variable $\delta_c$ represents the actuator command.

Since the standard air density model is a function of height $h$, the following complementary function is introduced
\begin{equation}
\dot h = V \sin \gamma
\label{eq:19}
\end{equation}

In autopilot design, the angle-of-attack $\alpha$ and the pitch rate $\dot \theta$ are considered as the state variables. The lateral acceleration $a_z$ is considered as the control output variable and the actuator command $\delta_c$ is regarded as the control input, that drives the lateral acceleration to track a reference command $a_{z,c}$.


\section{Computational Lateral Acceleration Autopilot}
\label{sec04}

\subsection{Learning Framework}

To apply the DDPG algorithm in autopilot design, we need to formulate the problem in the DRL framework. One intuitive choice is to utilise the entire dynamics model, detailed in Eqs. \eqref{eq:15}-\eqref{eq:19}, to represent the environment and directly learn the actuator command $\delta_c$ during agent training process. However, this simple learning procedure has been shown to be ineffective from our extensive test results. Our investigation suggests that it is mainly due to the nature of the vehicle dynamics: the considered longitudinal dynamics is statically unstable and therefore a small change in the control command could result in an unstable response. This hinders the learning effectiveness. Therefore, instead of learning the control action from scratch, we propose a new framework that utilises the domain knowledge to improve the learning effectiveness. To this end, we fix the autopilot structure with several feedback loops and leverage DDPG to learn the controller gains to implement the feedback controller. With this framework, it is also expected that the learning efficiency can be greatly improved. Notice that this concept is similar to fixed-structure $H_\infty$ control methodology \cite{apkarian2006nonsmooth,apkarian2015parametric}. However, our approach is a data-driven flight control algorithm and therefore has advantages against fixed-structure $H_\infty$ method.

\begin{figure*}[h]
\centering

\tikzset{every picture/.style={line width=0.75pt}}     

\begin{tikzpicture}[x=0.75pt,y=0.75pt,yscale=-1,xscale=1, scale=1]

\draw    (31.36,140.43) -- (176,140) ;
\draw   (162,89.31) .. controls (162,82.2) and (168.04,76.43) .. (175.5,76.43) .. controls (182.96,76.43) and (189,82.2) .. (189,89.31) .. controls (189,96.43) and (182.96,102.2) .. (175.5,102.2) .. controls (168.04,102.2) and (162,96.43) .. (162,89.31) -- cycle ; \draw   (165.95,80.2) -- (185.05,98.43) ; \draw   (185.05,80.2) -- (165.95,98.43) ;
\draw    (32.64,89.43) -- (80.5,90) ;
\draw    (175.5,102.2) -- (176,140) ;
\draw    (119,90.25) -- (162,90) ;
\draw   (232,73.4) -- (265,73.4) -- (265,106.4) -- (232,106.4) -- cycle ;
\draw    (30,190.5) -- (461,190) ;
\draw    (189,89.31) -- (232,89.06) ;
\draw    (265,89.31) -- (308,89.06) ;
\draw   (308,88.31) .. controls (308,81.2) and (314.04,75.43) .. (321.5,75.43) .. controls (328.96,75.43) and (335,81.2) .. (335,88.31) .. controls (335,95.43) and (328.96,101.2) .. (321.5,101.2) .. controls (314.04,101.2) and (308,95.43) .. (308,88.31) -- cycle ; \draw   (311.95,79.2) -- (331.05,97.43) ; \draw   (331.05,79.2) -- (311.95,97.43) ;
\draw    (335,89.31) -- (371,89) ;
\draw    (321.5,101.2) -- (322,190) ;
\draw   (371.5,62) -- (404,62) -- (404,117) -- (371.5,117) -- cycle ;
\draw    (404,89.2) -- (447,89.06) ;
\draw   (447,88.31) .. controls (447,81.2) and (453.04,75.43) .. (460.5,75.43) .. controls (467.96,75.43) and (474,81.2) .. (474,88.31) .. controls (474,95.43) and (467.96,101.2) .. (460.5,101.2) .. controls (453.04,101.2) and (447,95.43) .. (447,88.31) -- cycle ; \draw   (450.95,79.2) -- (470.05,97.43) ; \draw   (470.05,79.2) -- (450.95,97.43) ;
\draw   (517,73.4) -- (550,73.4) -- (550,106.4) -- (517,106.4) -- cycle ;
\draw    (474,89.31) -- (517,89.06) ;
\draw    (550,89.31) -- (593,89.06) ;
\draw   (80,76) -- (119.5,76) -- (119.5,103) -- (80,103) -- cycle ;
\draw    (460.5,101.2) -- (461,190) ;
\draw (168,89) node  [align=left] {{\footnotesize +}};
\draw (178,96) node  [align=left] {{\footnotesize -}};
\draw (99.75,89.5) node  [align=left] {$\displaystyle K_{DC}$};
\draw (248.5,89.9) node  [align=left] {$\displaystyle K_{A}$};
\draw (314,88) node  [align=left] {{\footnotesize +}};
\draw (323,95) node  [align=left] {\mbox{-}};
\draw (387.75,89.5) node  [align=left] {$\displaystyle \frac{K_{I}}{s}$};
\draw (453,88) node  [align=left] {{\footnotesize +}};
\draw (462,95) node  [align=left] {\mbox{-}};
\draw (533.5,89.9) node  [align=left] {$\displaystyle K_{g}$};
\draw (39,124) node  [align=left] {$\displaystyle a_{z}$};
\draw (40,77) node  [align=left] {$\displaystyle a_{z,c}$};
\draw (37,175) node  [align=left] {$\displaystyle q$};
\draw (583,75) node  [align=left] {$\displaystyle \delta _{c}$};

\end{tikzpicture}

\caption{Three-loop autopilot structure.}
\label{fig:three}
\end{figure*}
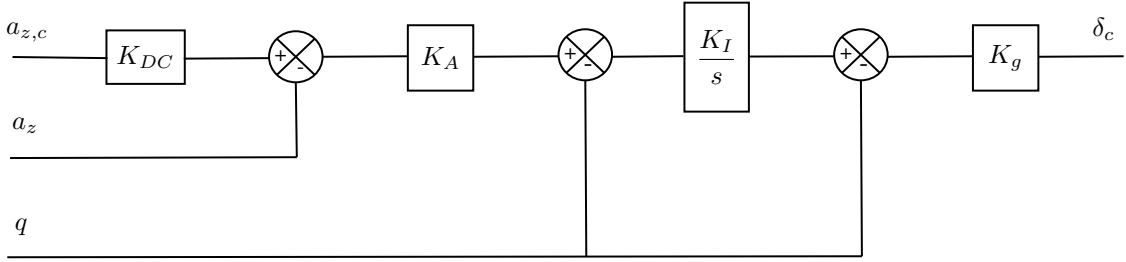

Over the past several decades, classical three-loop autopilot structures have been extensively employed for acceleration control for air vehicles due to its simple structure and effectiveness \cite{zarchan2012tactical,stein2016effect,mracek2005missile,lee2016connections,kim2017augmented}. The classical three-loop autopilot is given by a simple structure with two feedback loops, as shown in Fig. \ref{fig:three}. The inner loop utilises a proportional-integral feedback of pitch rate and the outer loop leverages proportional feedback of lateral acceleration. With this architecture, the autopilot gains $K_{DC}$, $K_A$, $K_I$ and $K_g$ are usually designed using linear control theory for several trim operation points individually. \cite{mracek2005missile} compared various three-loop topologies and showed that the gains can be optimally derived by using the LQR concepts. Note that the three-loop autopilot is realised by scheduling the gains with some external signals, e.g., angle-of-attack, Mach number, height in linear control. Due to this fact, implementing classical three-loop autopilot requires a look-up table and a proper scheduling algorithm. This fact inevitably increases the complexity of the controller and results in some approximation errors during the scheduling process. For modern air vehicles with large flight envelope, massive \textit{ad hoc} trim operation points are required to guarantee the performance of gain-scheduling. This further increases the complexity of autopilot design. 

In recent study \cite{lee2016connections}, there was an investigation to identify the connection between linear and nonlinear autopilots through three-loop topology. This study revealed that non-linear autopilot  shares the three-loop topology and the gains are parameter varying. The issue is that the performance of the non-linear controller can be significantly degraded with presence of system uncertainties, which could be the case in the modern air vehicles.  

This paper fixes the structure of the autopilot as three-loop topology. Note that we might be able to examine other autopilot topology to overcome the limitations of the conventional one and even the autopilot topology could be subject of learning itself. We will handle these points in our future study. 

To address the issues with conventional control theories discussed, this paper aims to utilise DDPG to provide a direct mapping from scheduling variables to autopilot gains, i.e.,
\begin{equation}
\begin{split}
& K_{DC}= f_1 \left(\alpha, M, h \right) \\ 
& K_{A}= f_2 \left(\alpha, M, h \right) \\ 
& K_{I}= f_3 \left(\alpha, M, h \right) \\ 
& K_{g}= f_4 \left(\alpha, M, h \right) \\ 
 \end{split}
\label{eq:20}
\end{equation}
where $f_i$ with $i=1,2,3,4$ are nonlinear functions.

In other words, we suggest to directly train a neural network that provides nonlinear transformations from scheduling variables $\alpha$, $M$, $h$, to autopilot gains $K_{DC}$, $K_A$, $K_I$, $K_g$.

\subsection{Problem Formulation}

To learn the autopilot gains using DDPG, we need to formulate the problem in the RL framework by constructing a MDP with a proper reward function. 

\subsubsection{MDP definition}
The dynamics of angle-of-attack, pitch angle, pitch rate, Mach number and height, shown in Eqs. \eqref{eq:15}-\eqref{eq:19}, constitutes the environment, which is fully characterised by the system state
\begin{equation}
s_t = \left(\alpha, q, \theta, M, h \right)
\label{eq:21}
\end{equation}
As stated before, the aim of DDPG here is to learn the autopilot gains. For this reason, the agent action is naturally defined as
\begin{equation}
a_t = \left(K_{DC}, K_A, K_I, K_g \right)
\label{eq:22a}
\end{equation}
From Eq. \eqref{eq:19}, it can be noted that the autopilot gains are functions of angle-of-attack, Mach number and height and these three variables are directly measurable from onboard sensors. For this reason, the agent observation is defined as
\begin{equation}
o_t = \left(\alpha, M, h \right)
\label{eq:22b}
\end{equation}
which gives a partially observable MDP. Note that the DDPG algorithm is applicable to partially observable MDP, as shown in \cite{duan2016benchmarking}. It is worthy pointing out that we can also include pitch angle and pitch rate in the observation vector during training. However, increasing the dimension of observation will increase the difficulty for the training process as more complicated network for function approximation is required. 

\begin{figure}[tb]
\centering

\tikzset{every picture/.style={line width=0.75pt}} 

\begin{tikzpicture}[x=0.75pt,y=0.75pt,yscale=-1,xscale=1]

\draw  [color={rgb, 255:red, 208; green, 2; blue, 27 }  ,draw opacity=1 ][line width=2.25]  (320,95.4) .. controls (320,89.1) and (325.1,84) .. (331.4,84) -- (444.1,84) .. controls (450.4,84) and (455.5,89.1) .. (455.5,95.4) -- (455.5,129.6) .. controls (455.5,135.9) and (450.4,141) .. (444.1,141) -- (331.4,141) .. controls (325.1,141) and (320,135.9) .. (320,129.6) -- cycle ;
\draw  [color={rgb, 255:red, 189; green, 16; blue, 224 }  ,draw opacity=1 ][line width=2.25]  (328,202.4) .. controls (328,196.1) and (333.1,191) .. (339.4,191) -- (434.1,191) .. controls (440.4,191) and (445.5,196.1) .. (445.5,202.4) -- (445.5,236.6) .. controls (445.5,242.9) and (440.4,248) .. (434.1,248) -- (339.4,248) .. controls (333.1,248) and (328,242.9) .. (328,236.6) -- cycle ;
\draw [color={rgb, 255:red, 208; green, 2; blue, 27 }  ,draw opacity=1 ][line width=2.25]    (387.5,190.38) -- (388.42,145) ;
\draw [shift={(388.5,141)}, rotate = 451.16] [color={rgb, 255:red, 208; green, 2; blue, 27 }  ,draw opacity=1 ][line width=2.25]    (17.49,-5.26) .. controls (11.12,-2.23) and (5.29,-0.48) .. (0,0) .. controls (5.29,0.48) and (11.12,2.23) .. (17.49,5.26)   ;

\draw [color={rgb, 255:red, 189; green, 16; blue, 224 }  ,draw opacity=1 ][line width=2.25]    (511.5,220) -- (449.5,220) ;
\draw [shift={(445.5,220)}, rotate = 360] [color={rgb, 255:red, 189; green, 16; blue, 224 }  ,draw opacity=1 ][line width=2.25]    (17.49,-5.26) .. controls (11.12,-2.23) and (5.29,-0.48) .. (0,0) .. controls (5.29,0.48) and (11.12,2.23) .. (17.49,5.26)   ;

\draw [color={rgb, 255:red, 189; green, 16; blue, 224 }  ,draw opacity=1 ][line width=2.25]    (513.5,112) -- (511.5,220) ;

\draw [color={rgb, 255:red, 208; green, 2; blue, 27 }  ,draw opacity=1 ][line width=2.25]    (253.5,220) -- (330.5,219) ;

\draw [color={rgb, 255:red, 208; green, 2; blue, 27 }  ,draw opacity=1 ][line width=2.25]    (253.5,111) -- (253.5,220) ;

\draw [color={rgb, 255:red, 208; green, 2; blue, 27 }  ,draw opacity=1 ][line width=2.25]    (253.5,111) -- (312.5,111) ;
\draw [shift={(316.5,111)}, rotate = 180] [color={rgb, 255:red, 208; green, 2; blue, 27 }  ,draw opacity=1 ][line width=2.25]    (17.49,-5.26) .. controls (11.12,-2.23) and (5.29,-0.48) .. (0,0) .. controls (5.29,0.48) and (11.12,2.23) .. (17.49,5.26)   ;

\draw [color={rgb, 255:red, 189; green, 16; blue, 224 }  ,draw opacity=1 ][line width=2.25]    (456.5,112) -- (513.5,112) ;

\draw (392.75,110.5) node [rotate=-0.8] [align=left] {\textbf{Agent (Flight }\\\textbf{Control System)}};
\draw (387.75,219.5) node [rotate=-0.8] [align=left] {\textbf{Flight Vehicle}};
\draw (423,166) node  [align=left] {\textbf{Reward}};
\draw (275,233) node  [align=left] {\textbf{Observation}};
\draw (486,235) node  [align=left] {\textbf{Action}};

\end{tikzpicture}

\caption{Information flow of the proposed RL framework.}
\label{fig:rl_concept}
\end{figure}
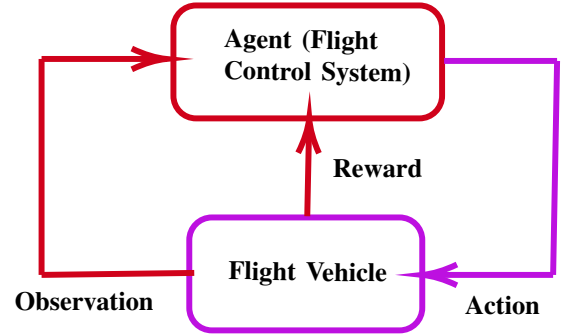

The relative kinematics \eqref{eq:15}-\eqref{eq:19}, environmental state \eqref{eq:21}, agent action \eqref{eq:22a}, agent observation \eqref{eq:22b}, together with a proper reward function, constitute a complete MDP formulation of the autopilot problem. The conceptual flowchart of the proposed flight control RL framework is shown in Fig. \ref{fig:rl_concept}.

\subsubsection{Reward function shaping}

The most challenging part of solving the autopilot design problem using DDPG is the development of a proper reward function. Notice that the primary objective of an acceleration autopilot is to drive the air vehicle to track a given acceleration command in a stable manner with acceptable transient as well as steady-state performance. In other words, the reward function should consider necessary time-domain metrics, e.g., rising time, overshoot, damping ratio, steady-state error, in an integrated manner. This means that the reward function should be designed as a weighted sum of several individual objectives, which, by default, poses great difficulty on tuning the weights of different metrics. This paper proposes to resolve this issue by utilising another domain knowledge, that is the desired performance. In other words, we shape the original reference command, $a_{z,c}$, based on the desired performance and propose to track the shaped command $\bar a_{z,c}$ rather than the original reference command. The shaped command $\bar a_{z,c}$ satisfies the following two properties:

(1) The shaped command is the output of a reference system;

(2) The reference system has desired time-domain characteristics.

This approach also enables alleviation of the particularity in applying RL/DRL to the control system design problem. One of the main objectives of the tracking control is to minimise the tracking error, that is the error between the reference command and actual output. If the tracking error is directly incorporated into the reward function with a discounting factor between 0 and 1, the learning-based control algorithm will consider minimising the immediate tracking error is more or equally important than minimising the tracking error in the future. This might cause instability issue and is not well aligned with the control design principles. Shaping the command and defining the tracking error with respect to the shaped command can relax this mismatch between the RL and control design concepts.

In consideration of the properties of a tail-controlled airframe, we propose the following reference system
\begin{equation}
\frac{{{{\bar a}_{z,c}}\left( s \right)}}{{{a_{z,c}}\left( s \right)}} = \frac{{ - 0.0363s + 1}}{{0.009{s^2} + 0.33s + 1}}
\label{eq:23}
\end{equation}
where the utilisation of an unstable zero naturally arises from the non-minimum phase property of a tail-controlled airframe.

The proposed reward function considers tradeoff between tracking error and fin deflection rate as
\begin{equation}
r_t =  - {k_a}{\left( {\frac{{{a_{z}} - {{\bar a}_{z,c}}}}{{{a_{z,\max }}}}} \right)^2} - {k_\delta }{\left( {\frac{{\dot \delta }}{{{{\dot \delta }_{\max }}}}} \right)^2}
\label{eq:24}
\end{equation}
where $k_a$ and $k_\delta$ are two positive constants that quantify the weights of two different objectives; $a_{z,\max}$ and ${\dot \delta}_{\max}$ are two normalisation constants that enforce these two metrics in approximately the same scale. Note that the consideration of fin deflection rate is to constrain the maximum rate of the actuator to meet physical limits. The hyper-parameters in shaping the reward function are summarised in Table \ref{tab:reward}.

\begin{table}[tb]
\begin{center}
\caption{Hyper-parameters in shaping the reward function.}
\label{tab:reward}
\begin{tabular}{cccc}
\hline\hline
$k_a$ & $k_\delta$ & $a_{z,\max}$ & ${\dot \delta}_{\max} $ \\ \hline
$1$ & $0.1$ & $100$ & $1.5$ \\
\hline\hline
\end{tabular}
\end{center}
\end{table}


\section{Training a DDPG Autopilot Agent}

Generally, training a DDPG agent involves three main steps: (1) obtaining training scenarios; (2) building the actor and critic networks; and (3) tuning the hyper parameters.

\subsubsection{Training scenarios}
In this paper, we consider an airframe with its flight envelop defined in Table \ref{tab:envelop}. At the beginning of each episode, we randomly initialise the system states with values uniformly distributed between the minimum and the maximum values. This random initialisation allows the agent to explore the diversity of the state space. For all episodes, the vehicle is required to track a reference command $a_{z,c}$, which is defined as a step command with its magnitude being $100m/s^2$.

\begin{table}[tb]
\begin{center}
\caption{Flight envelop.}
\label{tab:envelop}
\begin{tabular}{c|cc}
\hline\hline
Parameter & Minimum value & Maximum value \\ \hline
Angle-of-attack $\alpha$ & $-20^{\circ}$ & $20^{\circ}$ \\
Height $h$ & $6000m$ & $14000m$ \\
Mach number $M$  & $2$ & $4$ \\
\hline\hline
\end{tabular}
\end{center}
\end{table}

\subsubsection{Network construction}

\begin{table}[tb]
\begin{center}
\caption{Network layer size.}
\label{tab:nn}
\begin{tabular}{c|c|C{3cm}}
\hline\hline
Layer & Actor network & Critic network \\ \hline
Input layer& 3 (Size of observations) & 7 (Size of observations + Size of actions) \\
Hidden layer 1  & 64 & 64 \\
Hidden layer 2  & 64 & 64 \\
Output layer & 4 (Size of action) & 1 (Size of action-value function) \\
\hline\hline
\end{tabular}
\end{center}
\end{table}

Inspired by the original DDPG algorithm \cite{lillicrap2015continuous}, the actor and critic are represented by four-layer fully-connected neural networks. Note that this four-layer network architecture is commonly utilised in deep reinforcement learning applications \cite{henderson2018deep}. The layer sizes of these two networks are summarised in Table \ref{tab:nn}. Except for the actor output layer, each neuron in other layers is activated by a rectified linear units (Relu) function, which is defined as
\begin{equation} 
g(z)=\left\{\begin{array}{ll}{z,} & {\text { if } z>0} \\ {0,} & {\text { if } z<0}\end{array}\right.
\end{equation}
which provides faster processing speed than other nonlinear activation functions due to the linear relationship property.

The output layer of the actor network is activated by the $\tanh$ function, which is give by
\begin{equation} 
g(z)=\frac{e^{z}-e^{-z}}{e^{z}+e^{-z}}
\end{equation}
The benefit of the utilisation of $\tanh$ activation function in actor network is that it can prevent the control input from saturation as the actor output is constrained by $ \left(-1,1\right)$. Since different autopilot gains have different scales, the output layer of the actor network is scaled by a constant vector
\begin{equation} 
\left[K_{DC,\max}, K_{A,\max}, K_{I,\max}, K_{g,\max}\right]^T
\end{equation}
where $\left( \cdot \right)_{\max}$ stands for the normalisation constant of variable $\left( \cdot \right)$ and the detailed values are presented in Table \ref{tab:nor}.

As different observations have different scales and units, we normalise the observations at the input layers of the networks, thus providing unitless observations hat belong to approximately the same scale. This normalisation procedure is shown to be of paramount importance for our problems and helps to increase the training efficiency. Without normalisation, the average reward function cannot converge and even shows divergent patterns after some episodes. Denote $\bar {\left( \cdot \right)}$ as the normalised version of variable $\left( \cdot \right)$. The normalisation of observations is defined as
\begin{equation}
\bar {\alpha} = \frac{\alpha}{\alpha_{\max}}, \quad \bar {M} = \frac{M}{M_{\max}}, \quad \bar {h} = \frac{h}{h_{\max}} 
\end{equation}
where $\left( \cdot \right)_{\max}$ stands for the normalisation constant of variable $\left( \cdot \right)$ and the detailed values are presented in Table \ref{tab:nor}.

\begin{table}[tb]
\begin{center}
\caption{Normalisation constants.}
\label{tab:nor}
\begin{tabular}{C{0.5cm}C{0.5cm}C{0.5cm}C{0.9cm}C{0.8cm}C{0.8cm}C{0.8cm}}
\hline\hline
$\alpha_{\max}$ & $M_{\max}$ & $h_{\max}$ & $K_{DC,\max}$ & $K_{A,\max}$ & $K_{I,\max}$ & $K_{g,\max}$ \\ \hline
$20^{\circ}$ & 4 & 14000 & 3 & 0.05 & 100 & 2\\
\hline\hline
\end{tabular}
\end{center}
\end{table}

Both actor and critic networks are trained using Adam optimiser with $\mathcal L _2$ regularisation to address the over-fitting problem for stabilising the learning process. With $\mathcal L _2$ regularisation, the updates of actor and critic are modified as
\begin{equation} 
\begin{split}
\mathcal L _{actor} &= J\left( {{A^\mu }} \right) + \lambda_2 \mathcal L _2^A \\
\mu_{t+1} &= \mu_t +\alpha_\mu {\nabla _\mu }\mathcal L _{actor}
\end{split}
\end{equation}
\begin{equation}
\begin{split}
\mathcal L _{critic} &= \mathcal L(w) + \lambda_2 \mathcal L _2^C\\
w_{t+1} &= w_t +\alpha_w {\nabla _w }\mathcal L _{critic}
\end{split}
\end{equation}
where $\mathcal L _2^A$ and $\mathcal L _2^C$ denote the $\mathcal L _2$ regularisation losses on the weights of the actor and the critic, respectively; $\lambda_2$ is the regularisation constant.

To increase the stability of the network training process, we utilise the gradient clip technique to constrain the update of both actor and critic networks. More specifically, if the norm of the gradient exceeds a given upper bound $\rho$, the gradient is scaled to equal with $\rho$. This helps to prevent a numerical overflow or underflow during the training process.

\subsubsection{Hyper parameter tuning}

Each episode during training is terminated when the number of time steps exceeds the maximum permissible value. All hyper parameters that are utilised in DDPG training for our problem are summarised in Table \ref{tab:para}. Notice that the tuning of hyper parameters imposes great effects on the performance of DDPG and this tuning process is not consistent across different ranges of applications \cite{islam2017reproducibility,henderson2018deep}, i.e., different works utilised different set of hyper parameters for their own problems. For this reason, we tune these hyper parameters for our autopilot design problem based on several trial and error tests.


\begin{table}[tb]
\begin{center}
\caption{Hyper parameter settings.}
\label{tab:para}
\begin{tabular}{C{2.5cm}|c|C{2.5cm}|c}
\hline\hline
Parameter & Value & Parameter & Value \\ \hline
Maximum permissible steps  & 200 & Size of experience buffer $|\mathcal D|$ & $5\times 10^5$ \\
Maximum permissible episodes  & 1000 & Size of mini-batch samples $N$ & 64 \\
Actor learning rate $\alpha_\mu$ & $10^{-3}$ & Mean of exploration noise $\mu_v$ & 0 \\
Critic learning rate $\alpha_w$ & $10^{-3}$ & Initial variance of exploration noise $\Sigma_1$ & $0.1$ \\
$\mathcal L _2$ regularisation constant $\lambda_2$ & $6\times 10^{-3}$ & Variance decay rate $\epsilon$ & $10^{-6}$ \\
Gradient upper bound $\rho$ &1 & Mean attraction constant $\beta_{attract}$ & 0.15 \\
Discounting factor $\gamma$ & 0.99 & Target network smoother constant $\tau$ & 0.001 \\
Sampling time $T_s$ & $0.01s$ & & \\
\hline\hline
\end{tabular}
\end{center}
\end{table}

\section{Results}

\subsection{Training Results}

In order to demonstrate the importance of the utilisation of reference command and domain knowledge, we also carry out simulations without using the shaped reference command and domain knowledge, i.e., learning from scratch. Without the shaped reference command, the reward function becomes
\begin{equation}
r_t =  - {k_a}{\left( {\frac{{{a_{z}} - {{a}_{z,c}}}}{{{a_{z,\max }}}}} \right)^2} - {k_\delta }{\left( {\frac{{\dot \delta }}{{{{\dot \delta }_{\max }}}}} \right)^2}
\end{equation}

\begin{figure*}[h]
\centering
	\subfloat[With both domain knowledge and shaped reference command]{{\includegraphics[width=0.33\linewidth]{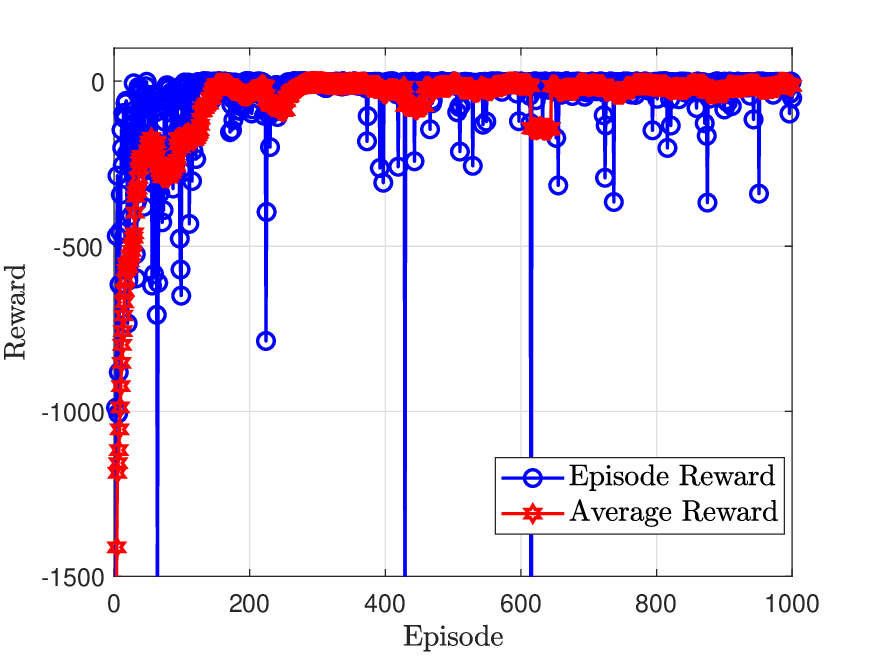}}}
	\subfloat[With only domain knowledge]{{\includegraphics[width=0.33\linewidth]{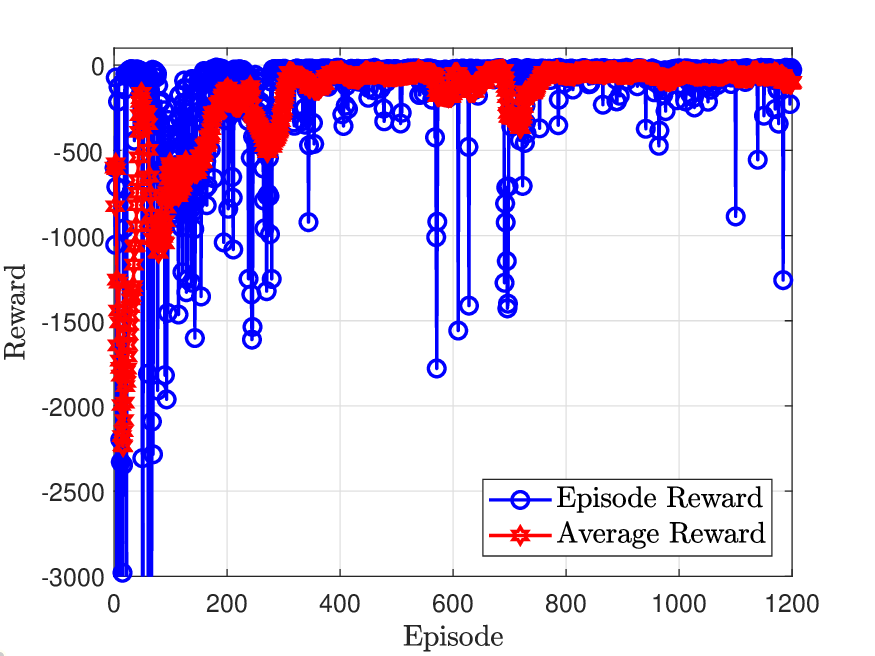}}}
	\subfloat[Learning from scratch for one fixed set point]{{\includegraphics[width=0.33\linewidth]{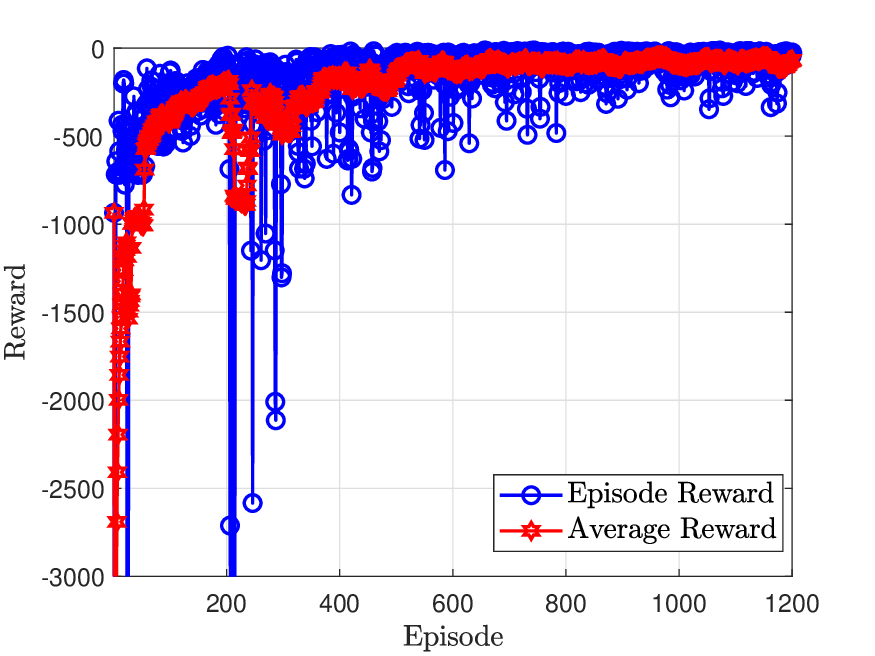}}}
\caption{Comparisons of learning curves.}
\label{fig:random_ini}
\end{figure*}

The learning curves of the training process that leverages domain knowledge with random initial conditions are shown in the first row of Fig. \ref{fig:random_ini}, where Fig. \ref{fig:random_ini} (a) is the result with shaped reference command and Fig. \ref{fig:random_ini} (b) stands for the result without shaped reference command. The average reward is obtained by averaging the episode reward within 30 episodes. From Figs. \ref{fig:random_ini} (a) and (b), it can be clearly noted that the average reward of the proposed DDPG autopilot agent converges to its steady-state value within 400 episodes, even with random initial conditions. Also, the utilisation of shaped reference command provides relatively faster convergence speed and smoother steady-state performance. To demonstrate this fact, Fig. \ref{fig:random_ini} (c) provides the learning curves of learning from scratch for one fixed set point. From this figure, it is clear that the reward convergence speed of learning from scratch for one fixed set point is slower than that of the proposed approach for all random initial conditions in the flight envelope. This fact reveals that the utilisation of domain knowledge significantly improves the learning efficiency. The reason is that the agent has already gained some experience by using the domain knowledge. 

\begin{figure}[ht]
\begin{center}
\includegraphics[width=0.5\textwidth]{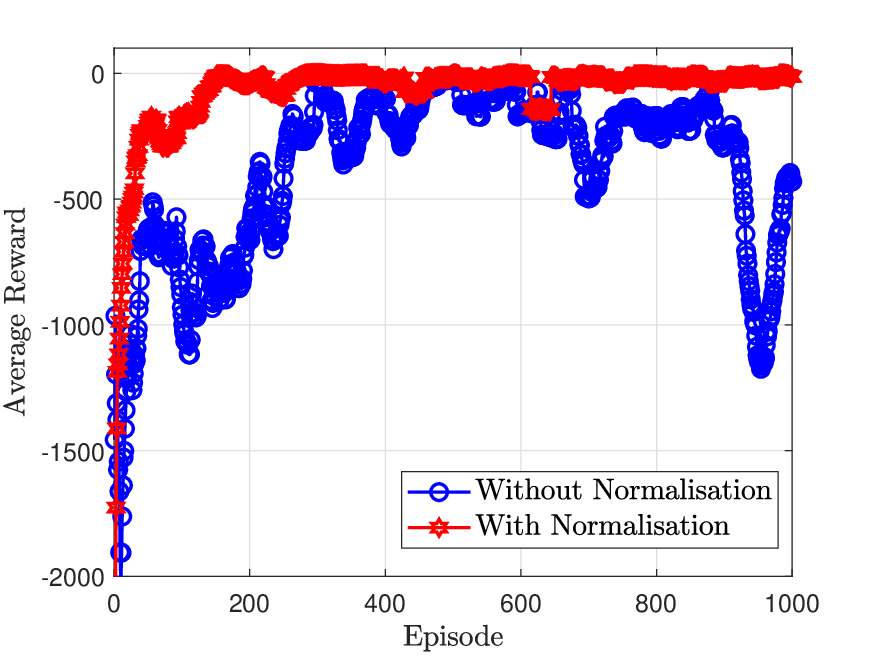}
\end{center}
    \caption{Learning process comparison with respect to normalisation.}
\label{fig:train_com}
\end{figure}

To investigate the importance of observation and action normalisation in training DDPG autopilot agent, we perform simulations with and without normalisation. Fig. \ref{fig:train_com} presents the comparison results of average reward convergence. From this figure, it is clear that utilising normalisation provides fast convergence rate of the learning process and higher steady-state value of the average reward function: the average reward without normalisation is not converged within 1,000 episodes, whereas the one with normalisation converges withing around 200 episodes. This means that leveraging observation and action normalisation helps to achieve more efficient and effective training process. The reason can be attributed to the fact that normalisation imposes equally importance on each element of the observation vector. Without normalisation, the scale difference between the elements varies in a great deal, e.g., the magnitude of height is much lager than that of angle-of-attack, and therefore prohibits effective training of the actor and critic networks. 

\subsection{Test Results}
\label{sub:test_result}
To test the proposed DDPG three-loop autopilot under various conditions, the trained agent is applied to some random scenarios and compared with classical gain scheduling approach. Note that the trained agent by learning from scratch is incapable to track reference signals that are different from the training process in our test. On the other hand, the trained agent by leveraging domain knowledge can track reference signals which differ from those in the training process. Therefore, we only compare the performance of the proposed algorithm with classical gain-scheduling and training with domain knowledge, but without shaped reference command. The testing DDPG agent is chosen from the one that generates the largest episode reward during the training process, i.e., -0.5527 (with the shaped reference command ) and -18.1761 (without the shaped reference command).

\begin{figure*}[h]
\centering
	\subfloat[Gain-scheduling]{{\includegraphics[width=0.33\linewidth]{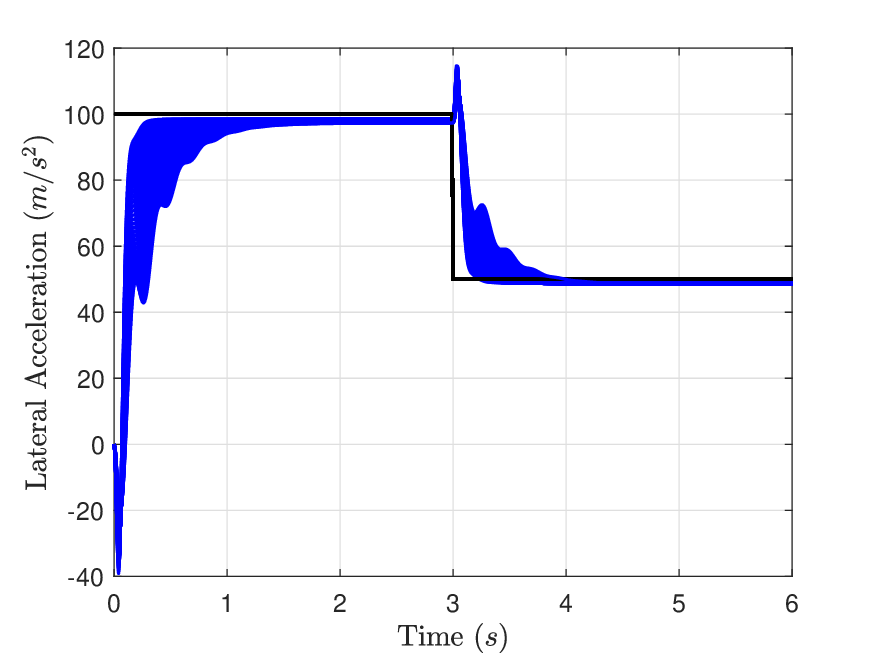}}}
	\subfloat[DDPG without shaped reference command]{{\includegraphics[width=0.33\linewidth]{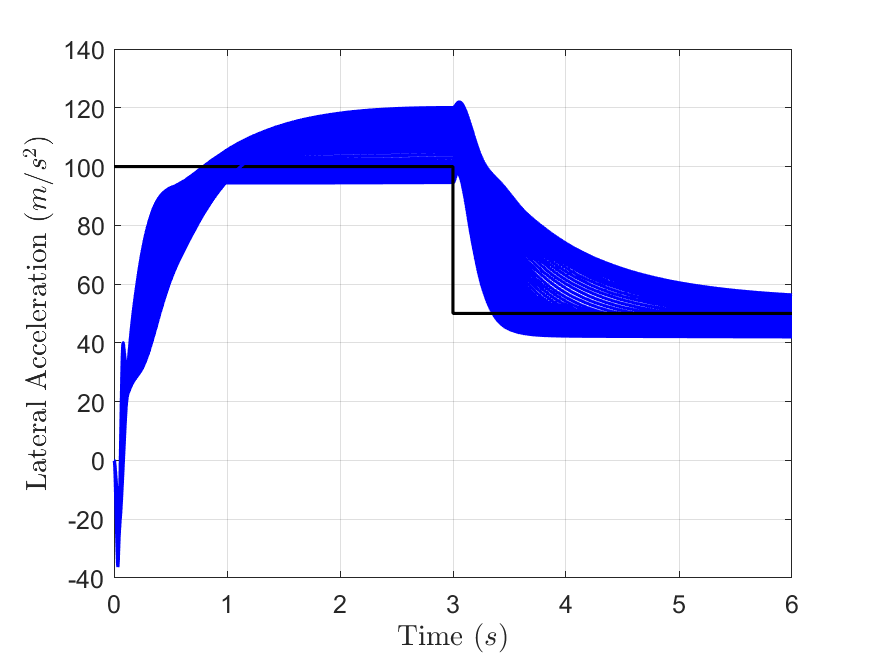}}}
	\subfloat[DDPG with shaped reference command]{{\includegraphics[width=0.33\linewidth]{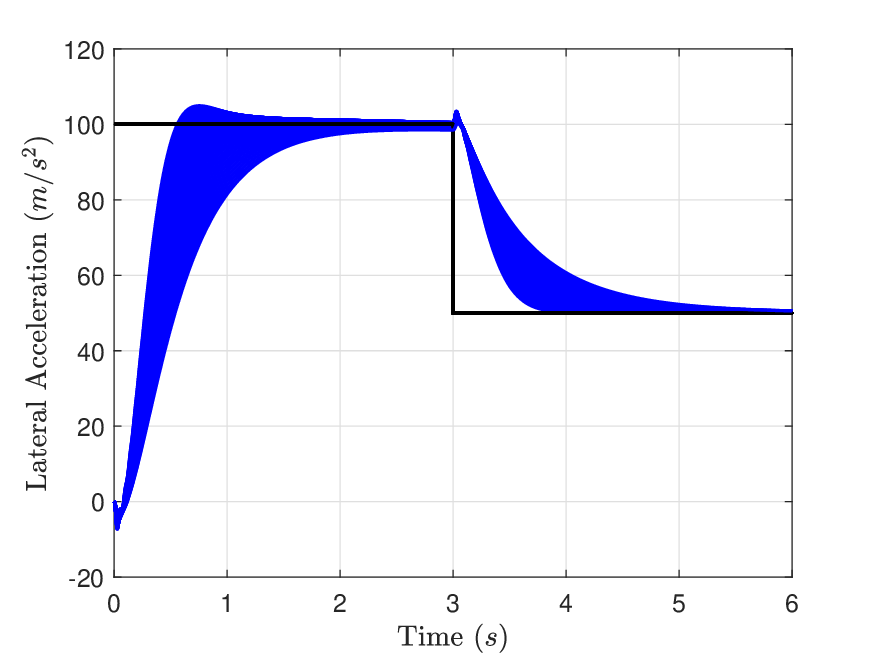}}}
    \caption{Comparison results of acceleration response.}
\label{fig:test1}
\end{figure*}

\begin{figure*}[h]
\centering
	\subfloat[Gain-scheduling]{{\includegraphics[width=0.33\linewidth]{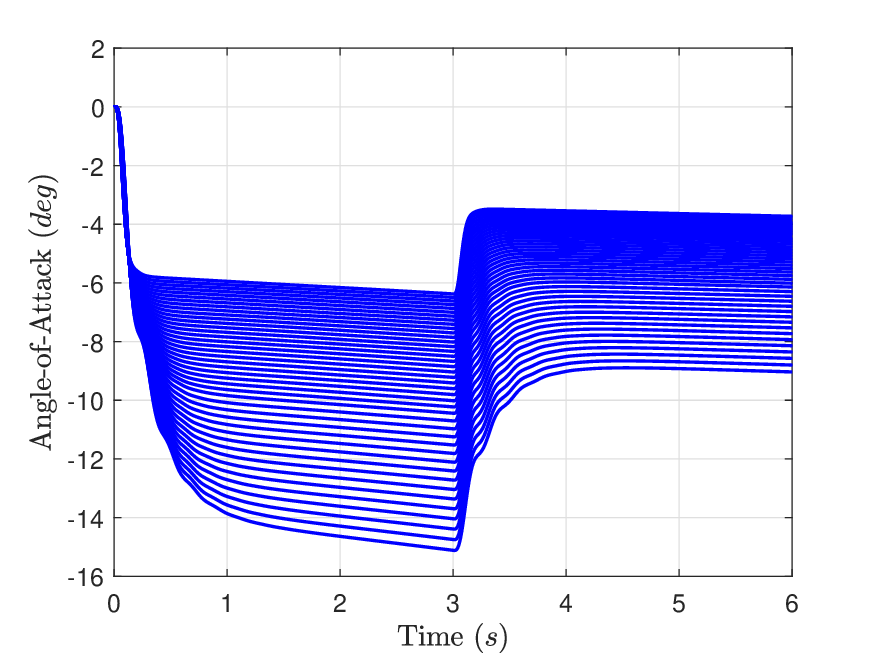}}}
	\subfloat[DDPG without shaped reference command]{{\includegraphics[width=0.33\linewidth]{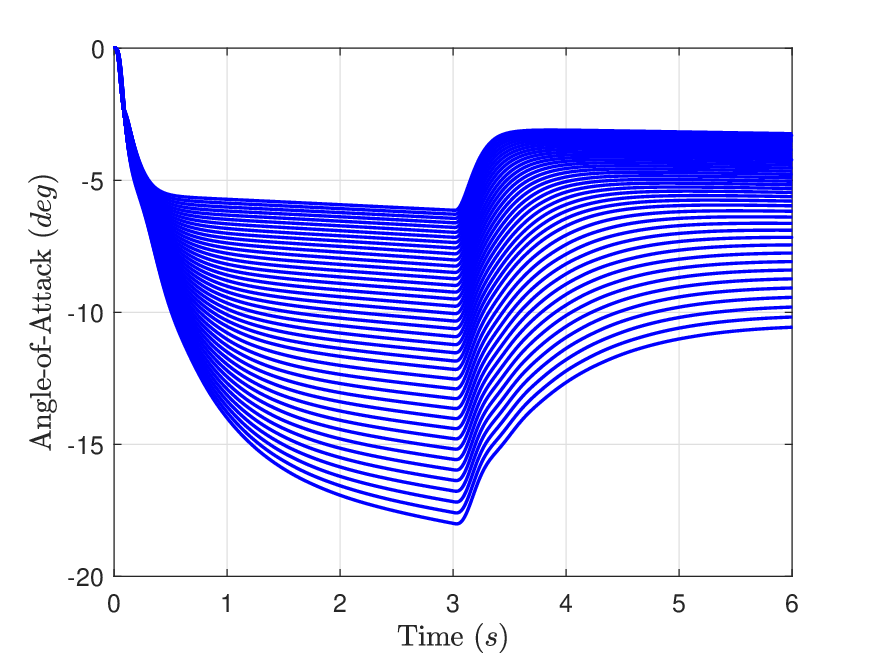}}} 
	\subfloat[DDPG with shaped reference command]{{\includegraphics[width=0.33\linewidth]{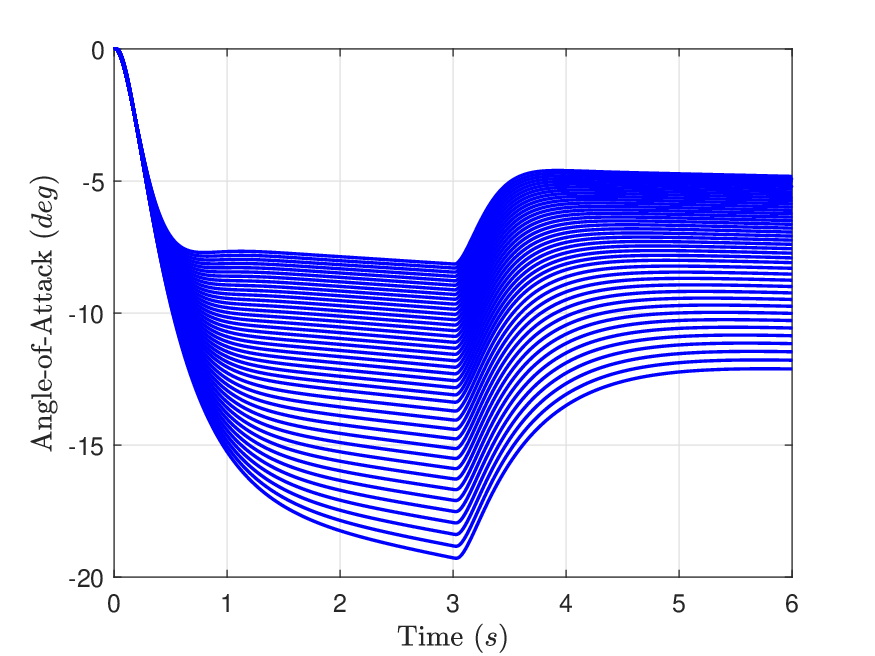}}}
    \caption{Comparison results of angle-of-attack.}
\label{fig:test2}
\end{figure*}

\begin{figure*}[h]
\centering
	\subfloat[Gain-scheduling]{{\includegraphics[width=0.33\linewidth]{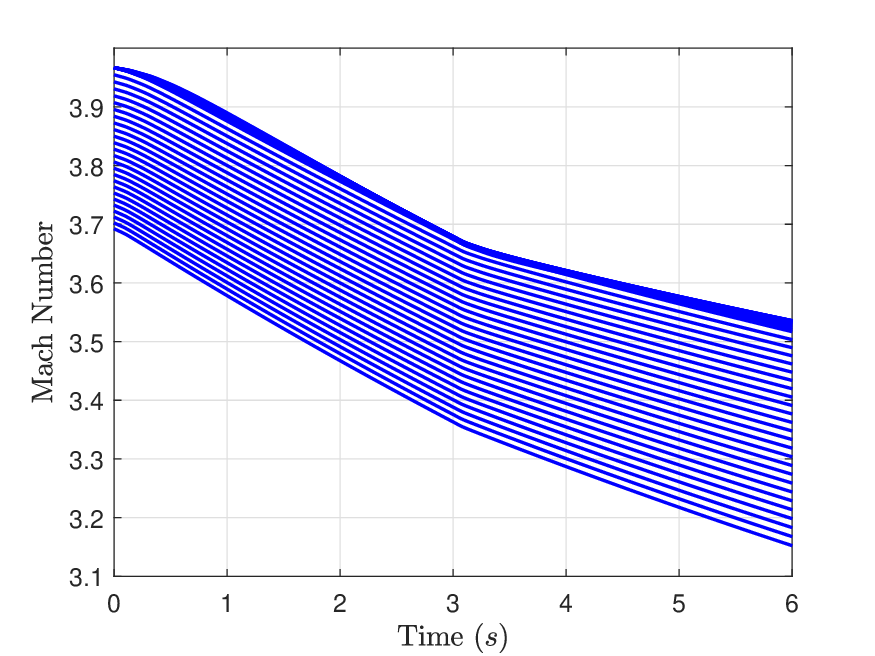}}}
	\subfloat[DDPG without shaped reference command]{{\includegraphics[width=0.33\linewidth]{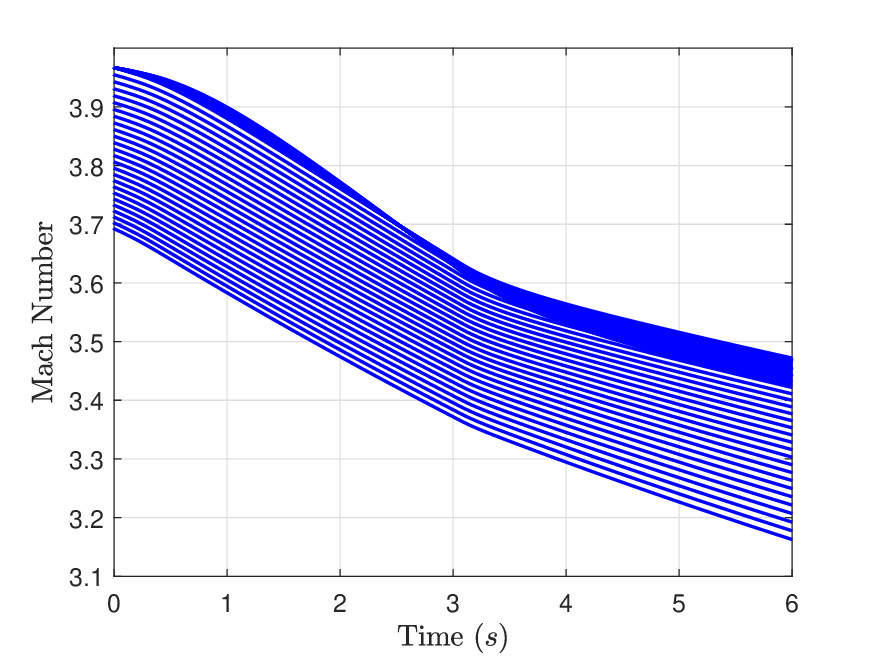}}} 
	\subfloat[DDPG with shaped reference command]{{\includegraphics[width=0.33\linewidth]{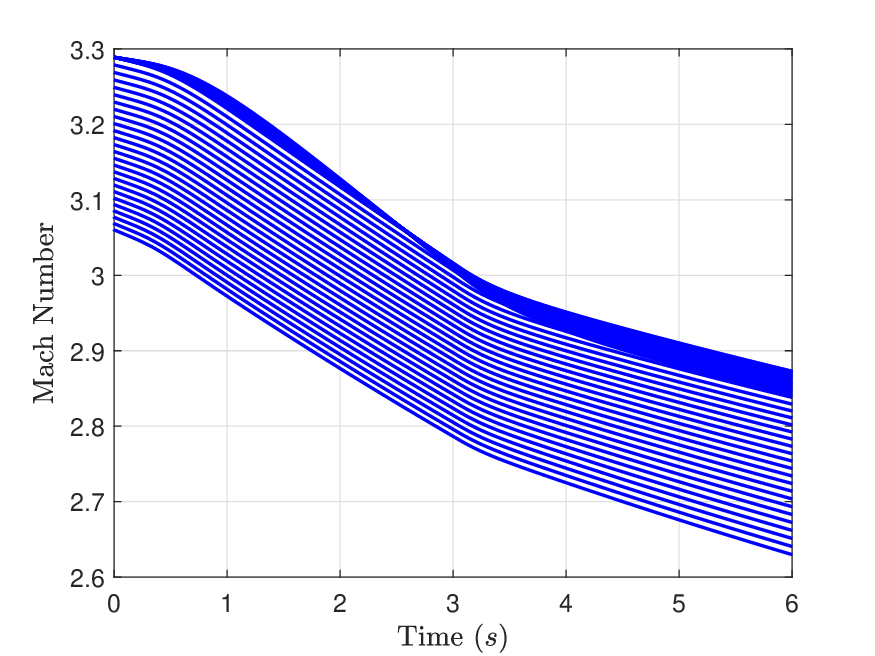}}}
    \caption{Comparison results of Mach number.}
\label{fig:test3}
\end{figure*}

\begin{figure*}[h]
\centering
	\subfloat[Gain-scheduling]{{\includegraphics[width=0.33\linewidth]{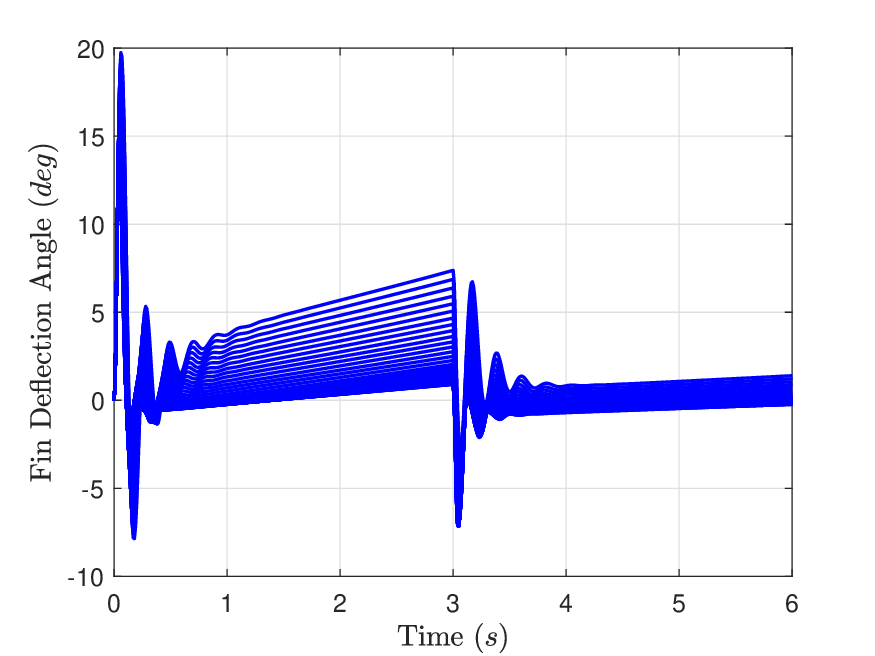}}}
	\subfloat[DDPG without shaped reference command]{{\includegraphics[width=0.33\linewidth]{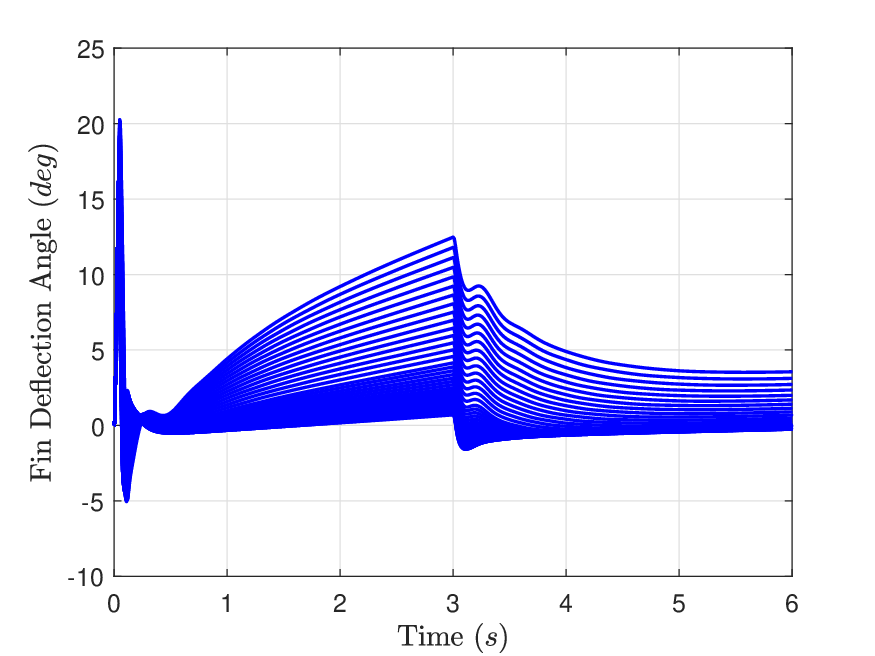}}} 
	\subfloat[DDPG with shaped reference command]{{\includegraphics[width=0.33\linewidth]{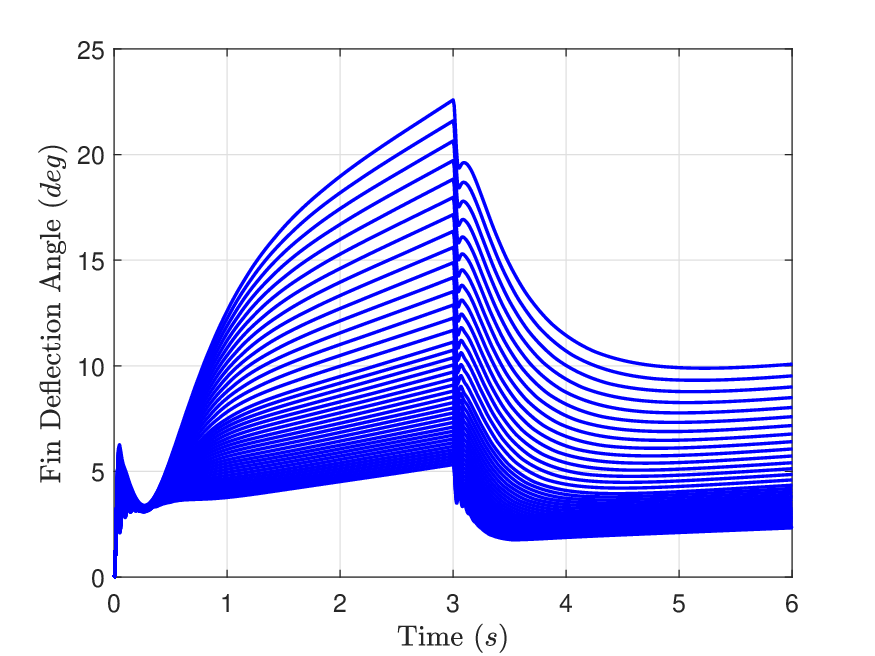}}}
    \caption{Comparison results of fin deflection angle.}
\label{fig:test4}
\end{figure*}

The comparison results, including acceleration response, angle-of-attack history, Mach number profile and fin deflection angle, are presented in Figs. \ref{fig:test1}-\ref{fig:test4}. From these figures, it can be observed that both the classical gain-scheduling and the proposed DDPG autopilots can somehow track the reference signal. Note that although the gains given a set point are designed to meet the typical autopilot design criteria for the gain scheduling, the set points are not optimally selected . Hence, response of the gain scheduling could be further improved by tuning the gains at more set points with more accurate model information. However, this is against with our arguments in Introduction and Section \ref{sec04}.  As a comparison, the agent that learned from data without using the shaped reference command provides poor transient performance, i.e., big undershoot at the beginning, and generates large performance variations even for the steady-state performance. Although the acceleration response under the proposed algorithm has larger overshoot than classical gain scheduling approach, the proposed autopilot shows more stable response with less response oscillations and smaller undershoot in all set points tested, as shown in Figs. \ref{fig:test1}. The test results reassure that leveraging domain knowledge can improve the learning effectiveness and generalisation.  

Fig. \ref{fig:test4} shows that gain scheduling autopilot requires faster actuator response and larger maximum fin deflection angle. This means that the proposed algorithm requires less actuator resource in tracking the reference command than gain scheduling approach. Another characteristic of the proposed autopilot is that it provides a direct nonlinear mapping from the scheduling variables to controller gains and therefore does not require a look-up table for real implementation. Apart from the improved learning effectiveness and generalisation, the numerical demonstration results suggest that the proposed DDPG three-loop autopilot provides comparable results with several advantages , compare to the classical gain scheduling approach, and could be a potential solution for three-loop autopilot design.

\subsection{Robustness Against Model Uncertainty}

\begin{figure*}[h]
\centering
	\subfloat[Acceleration response]{{\includegraphics[width=0.37\linewidth]{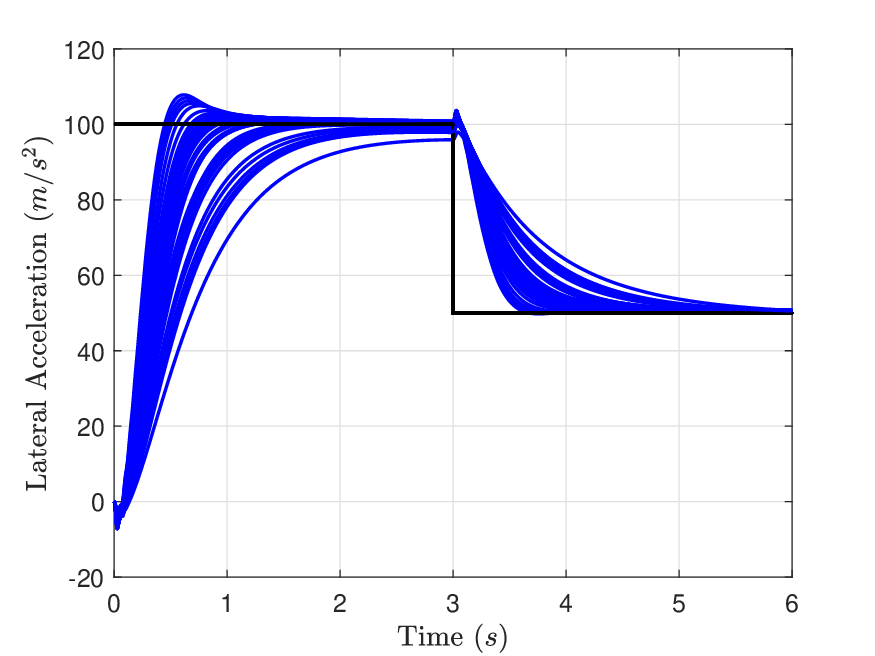}}}
	\subfloat[Angle-of-attack]{{\includegraphics[width=0.37\linewidth]{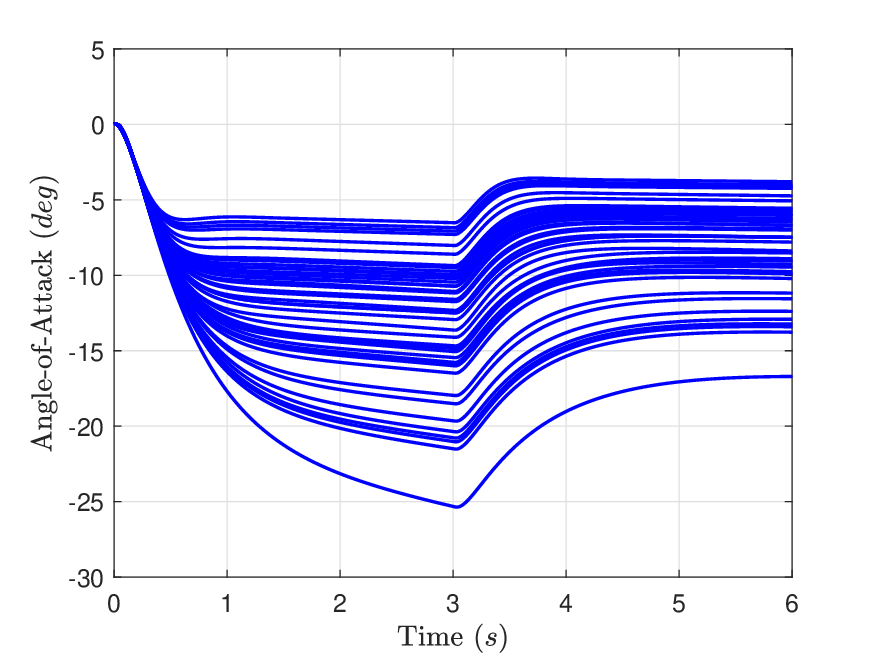}}} \\
	\subfloat[Mach number]{{\includegraphics[width=0.37\linewidth]{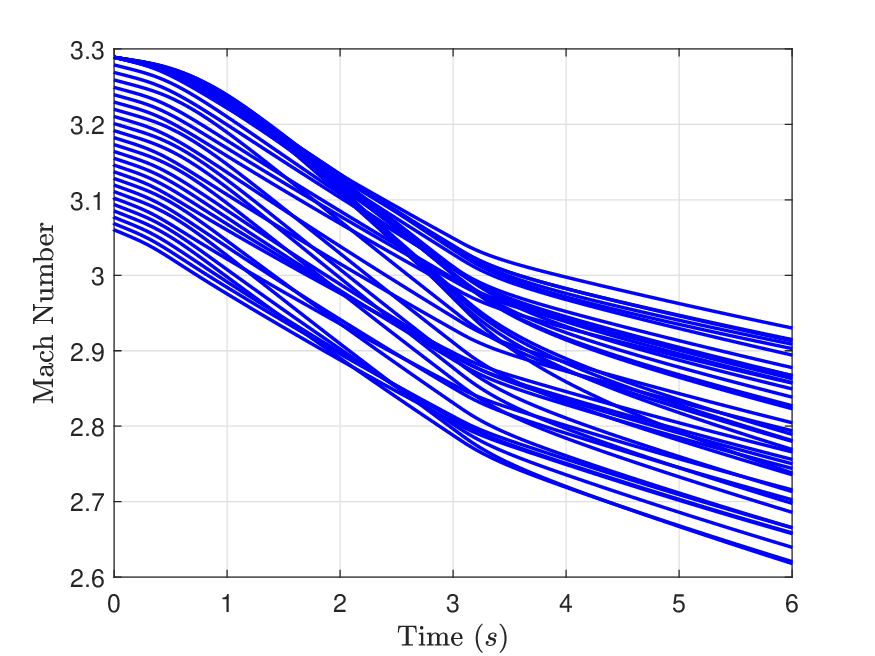}}}
	\subfloat[Fin deflection angle]{{\includegraphics[width=0.37\linewidth]{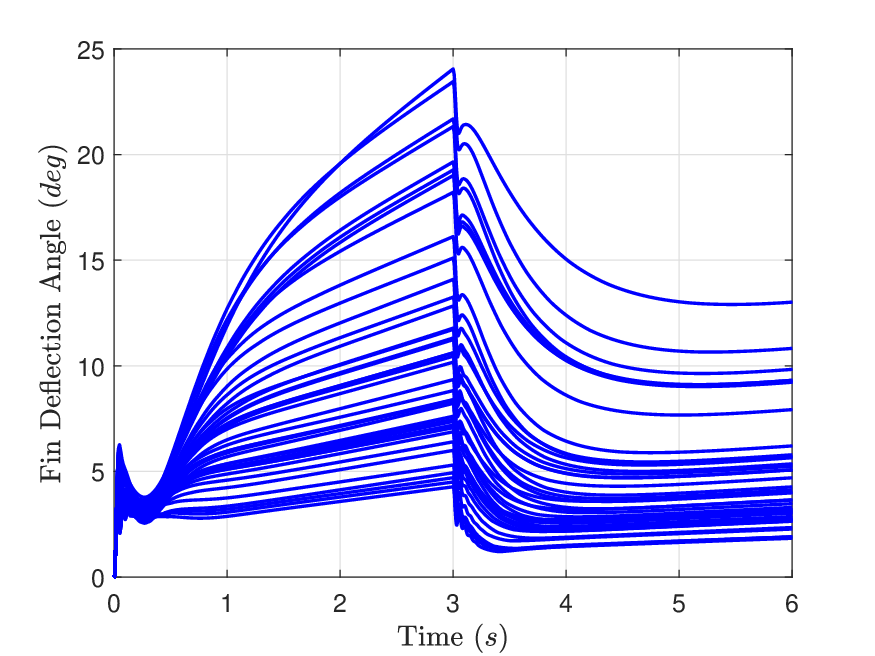}}}
    \caption{Simulation results of the proposed autopilot with model uncertainties.}
\label{fig:uncertain}
\end{figure*}

The aerodynamic model of a tail-controlled airframe might experience inevitable uncertainties due to the change of centre of pressure or other environmental parameters. Moreover, the data used in training could be different from reality. Therefore, it is critical to investigate the robustness of the learning-based autopilot.  To investigate the robustness of the proposed DRL-based acceleration autopilot, we perform numerical simulations with uncertain models in this subsection. The aerodynamic coefficients, detailed in Table \ref{tab:2}, are assumed to have random $-40\%$ to $+40\%$ uncertainty. To better show the robustness of the proposed control algorithm, these model uncertainties are not included in the training scenarios. The simulation results, including acceleration response, angle-of-attack, mach number and fin deflection angle, are presented in Fig. \ref{fig:uncertain}. From this figure, it can be clearly noted that the proposed DRL-based control algorithm provides satisfactory performance in the presence of aerodynamic model uncertainties: the model uncertainties only influence the transient effect and the steady-state tracking error remains at the same level as the case without aerodynamic uncertainty. These results clearly reveal that the proposed control approach has strong robustness against model uncertainties.

\subsection{Relative Stability}

\begin{figure}[h]
\centering

\tikzset{every picture/.style={line width=0.75pt}} 

\begin{tikzpicture}[x=0.75pt,y=0.75pt,yscale=-1,xscale=1,scale=.9]

\draw    (156.5,125) -- (181.5,125) ;
\draw [shift={(184.5,125)}, rotate = 180] [fill={rgb, 255:red, 0; green, 0; blue, 0 }  ][line width=0.08]  [draw opacity=0] (10.72,-5.15) -- (0,0) -- (10.72,5.15) -- (7.12,0) -- cycle    ;
\draw   (184,115.2) .. controls (184,111.78) and (186.78,109) .. (190.2,109) -- (252.3,109) .. controls (255.72,109) and (258.5,111.78) .. (258.5,115.2) -- (258.5,133.8) .. controls (258.5,137.22) and (255.72,140) .. (252.3,140) -- (190.2,140) .. controls (186.78,140) and (184,137.22) .. (184,133.8) -- cycle ;
\draw    (311.85,124.95) -- (336.85,124.95) ;
\draw [shift={(339.85,124.95)}, rotate = 180] [fill={rgb, 255:red, 0; green, 0; blue, 0 }  ][line width=0.08]  [draw opacity=0] (10.72,-5.15) -- (0,0) -- (10.72,5.15) -- (7.12,0) -- cycle    ;
\draw    (258.5,125) -- (283.5,125) ;
\draw [shift={(286.5,125)}, rotate = 180] [fill={rgb, 255:red, 0; green, 0; blue, 0 }  ][line width=0.08]  [draw opacity=0] (10.72,-5.15) -- (0,0) -- (10.72,5.15) -- (7.12,0) -- cycle    ;
\draw   (433,116.2) .. controls (433,112.78) and (435.78,110) .. (439.2,110) -- (492.3,110) .. controls (495.72,110) and (498.5,112.78) .. (498.5,116.2) -- (498.5,134.8) .. controls (498.5,138.22) and (495.72,141) .. (492.3,141) -- (439.2,141) .. controls (435.78,141) and (433,138.22) .. (433,134.8) -- cycle ;
\draw   (340,116.2) .. controls (340,112.78) and (342.78,110) .. (346.2,110) -- (398.3,110) .. controls (401.72,110) and (404.5,112.78) .. (404.5,116.2) -- (404.5,134.8) .. controls (404.5,138.22) and (401.72,141) .. (398.3,141) -- (346.2,141) .. controls (342.78,141) and (340,138.22) .. (340,134.8) -- cycle ;
\draw   (311.85,124.95) -- (286.46,138.11) -- (286.61,110.25) -- cycle ;
\draw    (404.85,125.95) -- (429.85,125.95) ;
\draw [shift={(432.85,125.95)}, rotate = 180] [fill={rgb, 255:red, 0; green, 0; blue, 0 }  ][line width=0.08]  [draw opacity=0] (10.72,-5.15) -- (0,0) -- (10.72,5.15) -- (7.12,0) -- cycle    ;
\draw    (498.85,125.95) -- (523.85,125.95) ;
\draw [shift={(526.85,125.95)}, rotate = 180] [fill={rgb, 255:red, 0; green, 0; blue, 0 }  ][line width=0.08]  [draw opacity=0] (10.72,-5.15) -- (0,0) -- (10.72,5.15) -- (7.12,0) -- cycle    ;
\draw   (318,175.2) .. controls (318,171.78) and (320.78,169) .. (324.2,169) -- (364.3,169) .. controls (367.72,169) and (370.5,171.78) .. (370.5,175.2) -- (370.5,193.8) .. controls (370.5,197.22) and (367.72,200) .. (364.3,200) -- (324.2,200) .. controls (320.78,200) and (318,197.22) .. (318,193.8) -- cycle ;
\draw    (219.5,185) -- (318.5,185) ;
\draw    (219.5,185) -- (219.5,143) ;
\draw [shift={(219.5,140)}, rotate = 450] [fill={rgb, 255:red, 0; green, 0; blue, 0 }  ][line width=0.08]  [draw opacity=0] (10.72,-5.15) -- (0,0) -- (10.72,5.15) -- (7.12,0) -- cycle    ;
\draw    (370.5,185) -- (468.5,185) ;
\draw    (468.5,185) -- (468.5,141) ;
\draw  [color={rgb, 255:red, 208; green, 2; blue, 27 }  ,draw opacity=1 ][dash pattern={on 4.5pt off 4.5pt}] (270,108.84) .. controls (270,103.41) and (274.41,99) .. (279.84,99) -- (310.66,99) .. controls (316.09,99) and (320.5,103.41) .. (320.5,108.84) -- (320.5,138.36) .. controls (320.5,143.79) and (316.09,148.2) .. (310.66,148.2) -- (279.84,148.2) .. controls (274.41,148.2) and (270,143.79) .. (270,138.36) -- cycle ;

\draw (154,101) node [anchor=north west][inner sep=0.75pt]   [align=left] {$\displaystyle a_{z,c}$};
\draw (288.22,116.97) node [anchor=north west][inner sep=0.75pt]   [align=left] {$\displaystyle k$};
\draw (505,100) node [anchor=north west][inner sep=0.75pt]   [align=left] {$\displaystyle a_{z}$};
\draw (261,72) node [anchor=north west][inner sep=0.75pt]   [align=left] {\textcolor[rgb]{0.82,0.01,0.11}{\textbf{Input gain}}};
\draw (190,116.2) node [anchor=north west][inner sep=0.75pt]   [align=left] {$\displaystyle \text{Autopilot}$};
\draw (321,176.2) node [anchor=north west][inner sep=0.75pt]   [align=left] {$\displaystyle \text{Sensor}$};
\draw (344,117.2) node [anchor=north west][inner sep=0.75pt]   [align=left] {$\displaystyle \text{Actuator}$};
\draw (435,117.2) node [anchor=north west][inner sep=0.75pt]   [align=left] {$\displaystyle \text{Airframe}$};

\end{tikzpicture}

\caption{Method to determine gain margin.}
\label{fig:input_gain}
\end{figure}
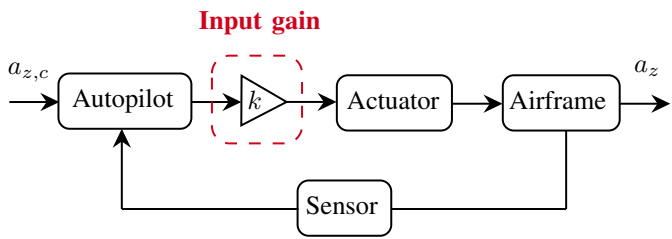

\begin{figure}[h]
\centering

\tikzset{every picture/.style={line width=0.75pt}} 

\begin{tikzpicture}[x=0.75pt,y=0.75pt,yscale=-1,xscale=1,scale=.9]

\draw    (156.5,125) -- (181.5,125) ;
\draw [shift={(184.5,125)}, rotate = 180] [fill={rgb, 255:red, 0; green, 0; blue, 0 }  ][line width=0.08]  [draw opacity=0] (10.72,-5.15) -- (0,0) -- (10.72,5.15) -- (7.12,0) -- cycle    ;
\draw   (184,115.2) .. controls (184,111.78) and (186.78,109) .. (190.2,109) -- (252.3,109) .. controls (255.72,109) and (258.5,111.78) .. (258.5,115.2) -- (258.5,133.8) .. controls (258.5,137.22) and (255.72,140) .. (252.3,140) -- (190.2,140) .. controls (186.78,140) and (184,137.22) .. (184,133.8) -- cycle ;
\draw    (258.5,125) -- (278.5,125) ;
\draw [shift={(281.5,125)}, rotate = 180] [fill={rgb, 255:red, 0; green, 0; blue, 0 }  ][line width=0.08]  [draw opacity=0] (10.72,-5.15) -- (0,0) -- (10.72,5.15) -- (7.12,0) -- cycle    ;
\draw   (433,116.2) .. controls (433,112.78) and (435.78,110) .. (439.2,110) -- (492.3,110) .. controls (495.72,110) and (498.5,112.78) .. (498.5,116.2) -- (498.5,134.8) .. controls (498.5,138.22) and (495.72,141) .. (492.3,141) -- (439.2,141) .. controls (435.78,141) and (433,138.22) .. (433,134.8) -- cycle ;
\draw   (346,116.2) .. controls (346,112.78) and (348.78,110) .. (352.2,110) -- (404.3,110) .. controls (407.72,110) and (410.5,112.78) .. (410.5,116.2) -- (410.5,134.8) .. controls (410.5,138.22) and (407.72,141) .. (404.3,141) -- (352.2,141) .. controls (348.78,141) and (346,138.22) .. (346,134.8) -- cycle ;
\draw   (323.41,125.37) -- (281.35,140.08) -- (281.52,108.77) -- cycle ;
\draw    (498.85,125.95) -- (523.85,125.95) ;
\draw [shift={(526.85,125.95)}, rotate = 180] [fill={rgb, 255:red, 0; green, 0; blue, 0 }  ][line width=0.08]  [draw opacity=0] (10.72,-5.15) -- (0,0) -- (10.72,5.15) -- (7.12,0) -- cycle    ;
\draw   (318,175.2) .. controls (318,171.78) and (320.78,169) .. (324.2,169) -- (364.3,169) .. controls (367.72,169) and (370.5,171.78) .. (370.5,175.2) -- (370.5,193.8) .. controls (370.5,197.22) and (367.72,200) .. (364.3,200) -- (324.2,200) .. controls (320.78,200) and (318,197.22) .. (318,193.8) -- cycle ;
\draw    (219.5,185) -- (318.5,185) ;
\draw    (219.5,185) -- (219.5,143) ;
\draw [shift={(219.5,140)}, rotate = 450] [fill={rgb, 255:red, 0; green, 0; blue, 0 }  ][line width=0.08]  [draw opacity=0] (10.72,-5.15) -- (0,0) -- (10.72,5.15) -- (7.12,0) -- cycle    ;
\draw    (370.5,185) -- (468.5,185) ;
\draw    (468.5,185) -- (468.5,141) ;
\draw  [color={rgb, 255:red, 208; green, 2; blue, 27 }  ,draw opacity=1 ][dash pattern={on 4.5pt off 4.5pt}] (268,112) .. controls (268,107.03) and (272.03,103) .. (277,103) -- (320.5,103) .. controls (325.47,103) and (329.5,107.03) .. (329.5,112) -- (329.5,139) .. controls (329.5,143.97) and (325.47,148) .. (320.5,148) -- (277,148) .. controls (272.03,148) and (268,143.97) .. (268,139) -- cycle ;
\draw    (323.41,125.37) -- (343.41,125.37) ;
\draw [shift={(346.41,125.37)}, rotate = 180] [fill={rgb, 255:red, 0; green, 0; blue, 0 }  ][line width=0.08]  [draw opacity=0] (10.72,-5.15) -- (0,0) -- (10.72,5.15) -- (7.12,0) -- cycle    ;
\draw    (410.5,126) -- (430.5,126) ;
\draw [shift={(433.5,126)}, rotate = 180] [fill={rgb, 255:red, 0; green, 0; blue, 0 }  ][line width=0.08]  [draw opacity=0] (10.72,-5.15) -- (0,0) -- (10.72,5.15) -- (7.12,0) -- cycle    ;

\draw (154,101) node [anchor=north west][inner sep=0.75pt]   [align=left] {$\displaystyle a_{z,c}$};
\draw (281.22,117.97) node [anchor=north west][inner sep=0.75pt]   [align=left] {$\displaystyle e^{-\Delta ts}$};
\draw (505,100) node [anchor=north west][inner sep=0.75pt]   [align=left] {$\displaystyle a_{z}$};
\draw (261,72) node [anchor=north west][inner sep=0.75pt]   [align=left] {\textcolor[rgb]{0.82,0.01,0.11}{\textbf{Time delay}}};
\draw (322,176.2) node [anchor=north west][inner sep=0.75pt]   [align=left] {$\displaystyle \text{Sensor}$};
\draw (437,118.2) node [anchor=north west][inner sep=0.75pt]   [align=left] {$\displaystyle \text{Airframe}$};
\draw (350,117.2) node [anchor=north west][inner sep=0.75pt]   [align=left] {$\displaystyle \text{Actuator}$};
\draw (191,116.2) node [anchor=north west][inner sep=0.75pt]   [align=left] {$\displaystyle \text{Autopilot}$};

\end{tikzpicture}

\caption{Method to determine phase margin.}
\label{fig:time_delay}
\end{figure}
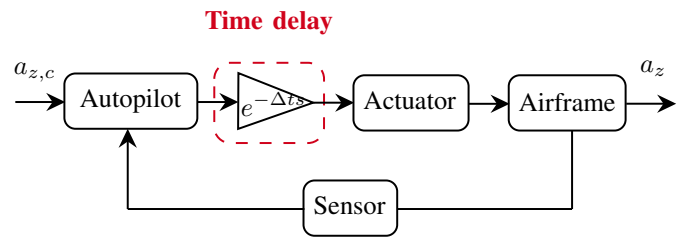

The relative stability is generally applied to examine the robustness of the control system within the linear control theory. However, since the proposed DRL autopilot is nonlinear and utilises the deep neural networks, it is hard to analyse the stability using classical linear control theory. To this end, this paper suggests to investigate relative stability of the proposed autopilot based on their physical concepts: (1) the gain margin determines the maximum allowable gain increase while maintaining the control loop stability; (2) the phase margin denotes the maximum allowable time delay in the control loop to guarantee stability. With these concepts in mind, the gain margin can by numerically computed as follows:

\begin{enumerate}
\item Put an input gain $k$ before the actuator, as shown in Fig. \ref{fig:input_gain}, and find the maximum value of $k$ that results in instability of the control loop.
\item The gain margin can then be determined as
\end{enumerate}
\begin{equation}
\text{GM}=20\log (k)\quad \left(\text{dB}\right)
\end{equation}

The phase margin can be also numerically determined as:
\begin{enumerate}
\item Put a time delay $\Delta t$ before the actuator, as shown in Fig. \ref{fig:time_delay}, and find the maximum value of $\Delta t$ that results in instability of the control loop.
\item Compute the gain crossover frequency $f$ from the obtained time response (the frequency can be directly read from the oscillation response).
\item The phase margin can then be determined as
\end{enumerate}
\begin{equation}
\text{PM}=360f\Delta t\quad \left(\text{deg}\right)
\end{equation}

\begin{figure*}[h]
\centering
	\subfloat[$h=6km$]{{\includegraphics[width=0.33\linewidth]{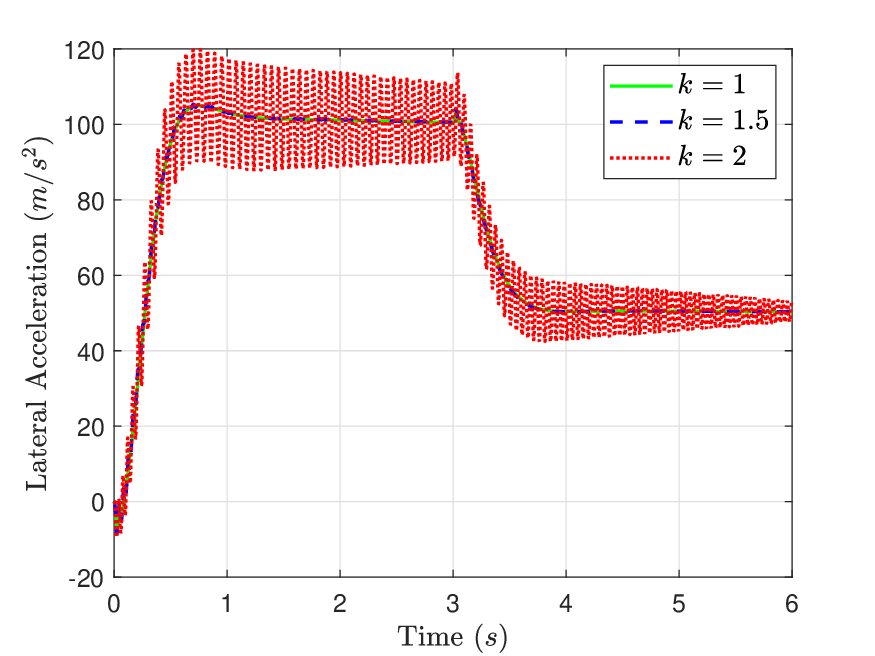}}}
	\subfloat[$h=8km$]{{\includegraphics[width=0.33\linewidth]{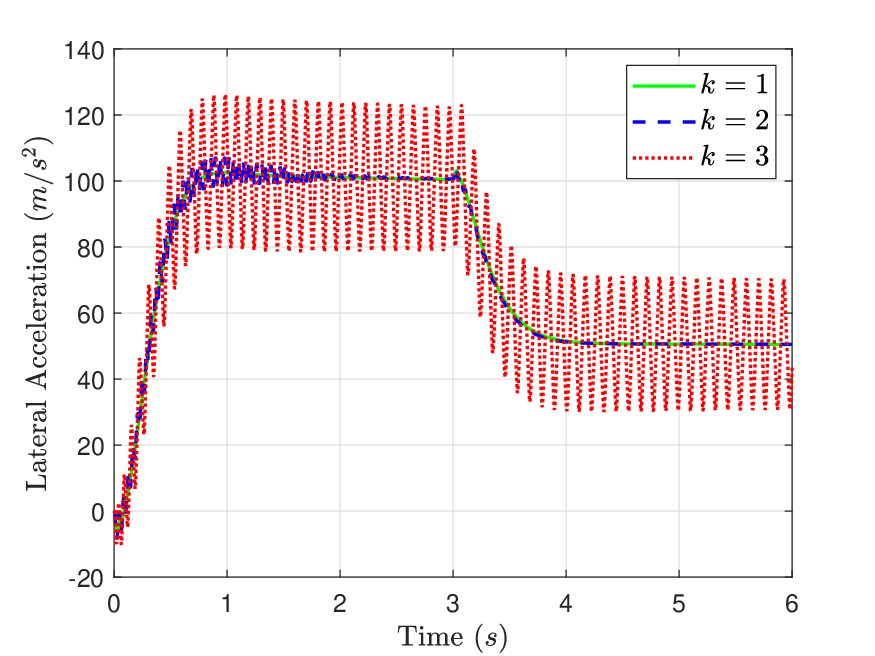}}}
	\subfloat[$h=10km$]{{\includegraphics[width=0.33\linewidth]{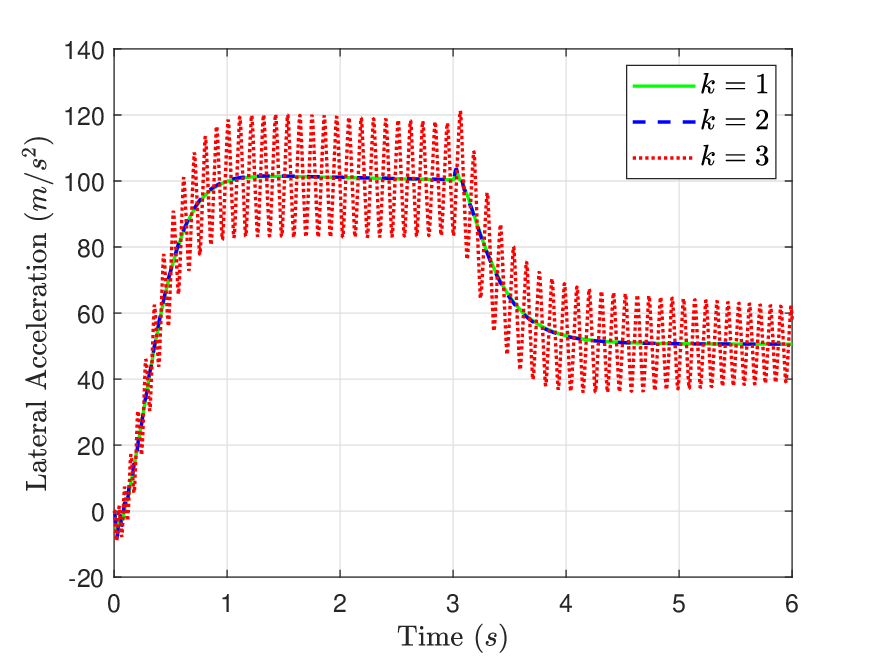}}}
    \caption{Relative stability analysis with gain margin.}
\label{fig:gain_margin}
\end{figure*}

\begin{figure*}[h]
\centering
	\subfloat[$h=6km$]{{\includegraphics[width=0.33\linewidth]{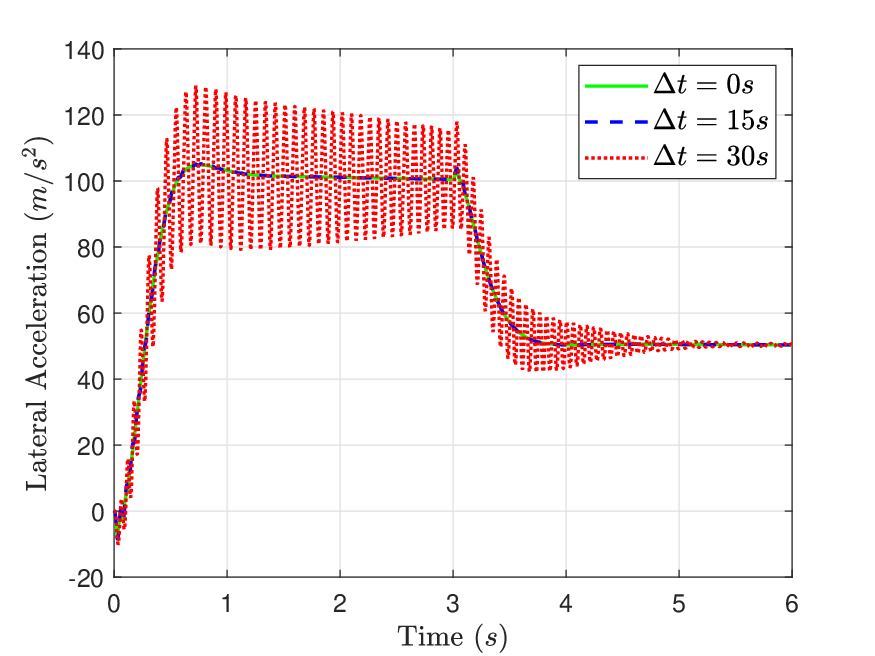}}}
	\subfloat[$h=8km$]{{\includegraphics[width=0.33\linewidth]{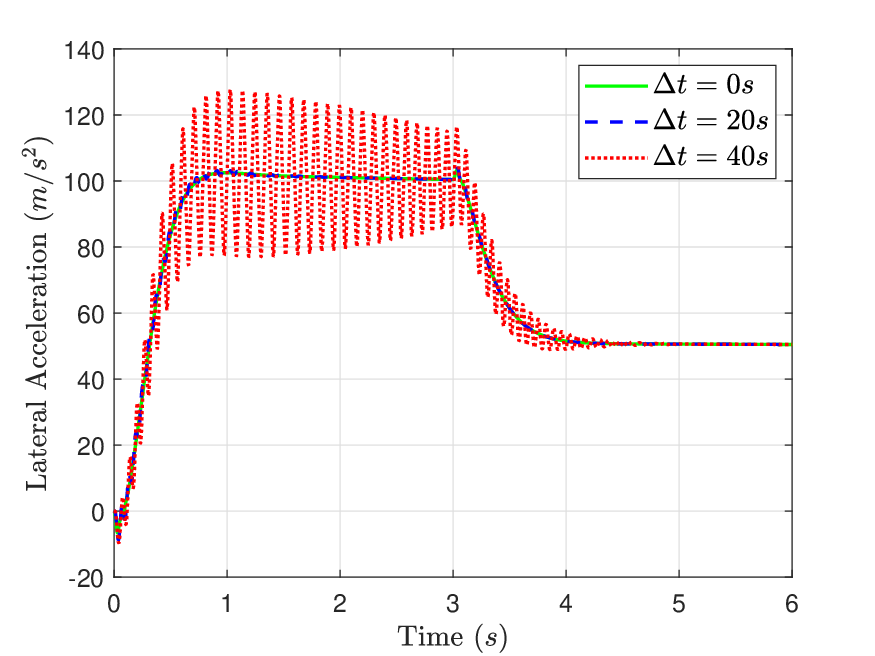}}}
	\subfloat[$h=10km$]{{\includegraphics[width=0.33\linewidth]{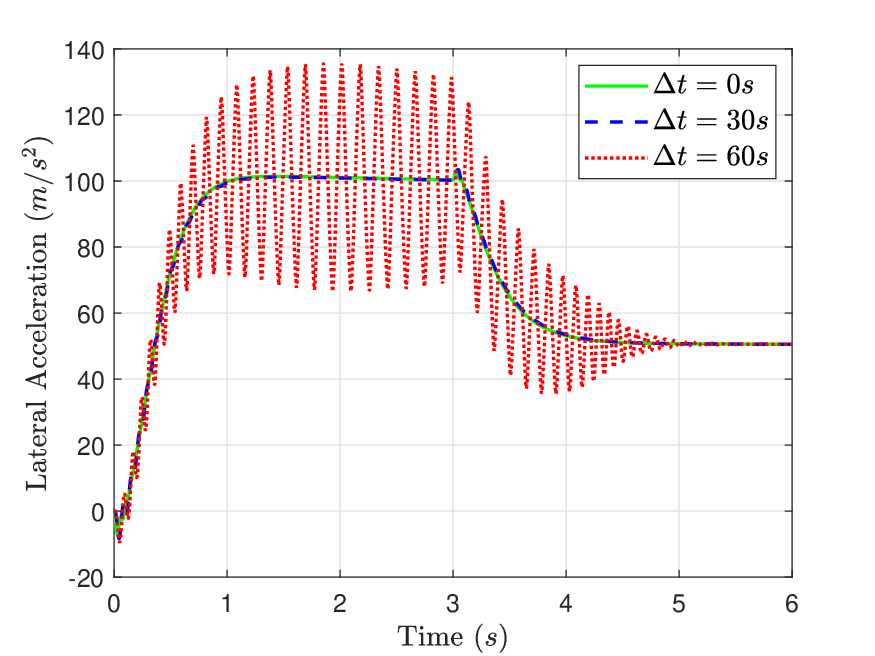}}}
    \caption{Relative stability analysis with phase margin.}
\label{fig:phase_margin}
\end{figure*}

We investigated relative stability in various set points and their the results are similar. Hence, we select three different heights: $6km$, $8km$ and $10km$ to demonstrate the results of relative stability examination. Notice that when performing relative stability analysis, the Mach number is time-varying. The numerical tests of gain margin and phase margin are presented Figs. \ref{fig:gain_margin} and \ref{fig:phase_margin}. From the results, we can readily obtain the result that the proposed autopilot satisfies typical design criteria:

\[
\text{GM}>6\text{dB}, \quad \text{PM}>45^{\circ}
\]

\section{Conclusion}

We have developed a computational acceleration autopilot algorithm for a tail-controlled air vehicle using deep RL techniques. The domain knowledge is utilised to help increase the learning efficiency during the training process. The state-of-the-art DDPG approach is leveraged to train a RL agent with a deterministic action policy that maximises the expected total reward. Extensive numerical simulations validate the effectiveness of the proposed approach. Future work includes extending the proposed autopilot to other types of vehicles. Validating the proposed computational autopilot algorithm under uncertain environment is also an important issue and requires further explorations.



\bibliographystyle{IEEEtran}
\bibliography{DDPG_Autopilot}

\begin{thebibliography}{10}
\providecommand{\url}[1]{#1}
\csname url@samestyle\endcsname
\providecommand{\newblock}{\relax}
\providecommand{\bibinfo}[2]{#2}
\providecommand{\BIBentrySTDinterwordspacing}{\spaceskip=0pt\relax}
\providecommand{\BIBentryALTinterwordstretchfactor}{4}
\providecommand{\BIBentryALTinterwordspacing}{\spaceskip=\fontdimen2\font plus
\BIBentryALTinterwordstretchfactor\fontdimen3\font minus
  \fontdimen4\font\relax}
\providecommand{\BIBforeignlanguage}[2]{{%
\expandafter\ifx\csname l@#1\endcsname\relax
\typeout{** WARNING: IEEEtran.bst: No hyphenation pattern has been}%
\typeout{** loaded for the language `#1'. Using the pattern for}%
\typeout{** the default language instead.}%
\else
\language=\csname l@#1\endcsname
\fi
#2}}
\providecommand{\BIBdecl}{\relax}
\BIBdecl

\bibitem{zarchan2012tactical}
P.~Zarchan, \emph{Tactical and strategic missile guidance}.\hskip 1em plus
  0.5em minus 0.4em\relax American Institute of Aeronautics and Astronautics,
  2012.

\bibitem{stilwell2001state}
D.~J. Stilwell, ``State-space interpolation for a gain-scheduled autopilot,''
  \emph{Journal of Guidance, Control, and Dynamics}, vol.~24, no.~3, pp.
  460--465, 2001.

\bibitem{stilwell1999interpolation}
D.~J. Stilwell and W.~J. Rugh, ``Interpolation of observer state feedback
  controllers for gain scheduling,'' \emph{IEEE Transactions on Automatic
  Control}, vol.~44, no.~6, pp. 1225--1229, 1999.

\bibitem{theodoulis2009missile}
S.~Theodoulis and G.~Duc, ``Missile autopilot design: gain-scheduling and the
  gap metric,'' \emph{Journal of Guidance, Control, and Dynamics}, vol.~32,
  no.~3, pp. 986--996, 2009.

\bibitem{lhachemi2016gain}
H.~Lhachemi, D.~Saussi{\'e}, and G.~Zhu, ``Gain-scheduling control design in
  the presence of hidden coupling terms,'' \emph{Journal of Guidance, Control,
  and Dynamics}, pp. 1872--1880, 2016.

\bibitem{thukral1998sliding}
A.~Thukral and M.~Innocenti, ``A sliding mode missile pitch autopilot synthesis
  for high angle of attack maneuvering,'' \emph{IEEE Transactions on Control
  Systems Technology}, vol.~6, no.~3, pp. 359--371, 1998.

\bibitem{shkolnikov2000robust}
I.~Shkolnikov, Y.~Shtessel, D.~Lianos, and A.~Thies, ``Robust missile autopilot
  design via high-order sliding mode control,'' in \emph{AIAA Guidance,
  navigation, and control Conference and Exhibit}, 2000.

\bibitem{mattei2014nonlinear}
G.~Mattei and S.~Monaco, ``Nonlinear autopilot design for an asymmetric missile
  using robust backstepping control,'' \emph{Journal of Guidance, Control, and
  Dynamics}, vol.~37, no.~5, pp. 1462--1476, 2014.

\bibitem{calise2000adaptive}
A.~J. Calise, M.~Sharma, and J.~E. Corban, ``Adaptive autopilot design for
  guided munitions,'' \emph{Journal of Guidance, Control, and Dynamics},
  vol.~23, no.~5, pp. 837--843, 2000.

\bibitem{wang2008l1}
J.~Wang, C.~Cao, N.~Hovakimyan, R.~Hindman, and D.~B. Ridgely, ``{$L_1$}
  adaptive controller for a missile longitudinal autopilot design,'' in
  \emph{AIAA Guidance, Navigation and Control Conference and Exhibit}, 2008, p.
  6282.

\bibitem{mracek1997full}
C.~Mracek, J.~Cloutier, J.~Cloutier, and C.~Mracek, ``Full envelope missile
  longitudinal autopilot design using the state-dependent riccati equation
  method,'' in \emph{Guidance, Navigation, and Control Conference}, 1997.

\bibitem{mracek2007sdre}
C.~P. Mracek, ``{SDRE} autopilot for dual controlled missiles,'' \emph{IFAC
  Proceedings Volumes}, vol.~40, no.~7, pp. 750--755, 2007.

\bibitem{buschek2003design}
H.~Buschek, ``Design and flight test of a robust autopilot for the iris-t
  air-to-air missile,'' \emph{Control Engineering Practice}, vol.~11, no.~5,
  pp. 551--558, 2003.

\bibitem{kim2017augmented}
J.-H. Kim and I.~H. Whang, ``Augmented three-loop autopilot structure based on
  mixed-sensitivity {$H_\infty$} optimization,'' \emph{Journal of Guidance,
  Control, and Dynamics}, vol.~41, no.~3, pp. 751--756, 2017.

\bibitem{lee2016connections}
C.-H. Lee, B.-E. Jun, and J.-I. Lee, ``Connections between linear and nonlinear
  missile autopilots via three-loop topology,'' \emph{Journal of Guidance,
  Control, and Dynamics}, vol.~39, no.~6, pp. 1426--1432, 2016.

\bibitem{lu2017introducing}
P.~Lu, ``Introducing computational guidance and control,'' \emph{Journal of
  Guidance, Control, and Dynamics}, vol.~40, no.~2, pp. 193--193, 2017.

\bibitem{tang2012predictive}
W.-Q. Tang and Y.-L. Cai, ``Predictive functional control-based missile
  autopilot design,'' \emph{Journal of Guidance, Control, and Dynamics},
  vol.~35, no.~5, pp. 1450--1455, 2012.

\bibitem{bachtiar2014nonlinear}
V.~Bachtiar, T.~M{\"u}hlpfordt, W.~Moase, T.~Faulwasser, R.~Findeisen, and
  C.~Manzie, ``Nonlinear model predictive missile control with a stabilising
  terminal constraint,'' \emph{IFAC Proceedings Volumes}, vol.~47, no.~3, pp.
  457--462, 2014.

\bibitem{bachtiar2017nonlinear}
V.~Bachtiar, C.~Manzie, and E.~C. Kerrigan, ``Nonlinear model-predictive
  integrated missile control and its multiobjective tuning,'' \emph{Journal of
  Guidance, Control, and Dynamics}, vol.~40, no.~11, pp. 2961--2970, 2017.

\bibitem{krishnakumar1992control}
K.~Krishnakumar and D.~E. Goldberg, ``Control system optimization using genetic
  algorithms,'' \emph{Journal of Guidance, Control, and Dynamics}, vol.~15,
  no.~3, pp. 735--740, 1992.

\bibitem{karimi2011multivariable}
B.~Karimi, I.~Saboori, and M.~Lotfi-Forushani, ``Multivariable controller
  design for aircraft longitudinal autopilot based on particle swarm
  optimization algorithm,'' in \emph{2011 IEEE International Conference on
  Computational Intelligence for Measurement Systems and Applications (CIMSA)
  Proceedings}.\hskip 1em plus 0.5em minus 0.4em\relax IEEE, 2011, pp. 1--6.

\bibitem{wang2017adaptive}
D.~Wang, H.~He, and D.~Liu, ``Adaptive critic nonlinear robust control: A
  survey,'' \emph{IEEE Transactions on Cybernetics}, vol.~47, no.~10, pp.
  3429--3451, 2017.

\bibitem{li2019transforming}
Y.~Li, Y.~Wen, K.~Guan, and D.~Tao, ``Transforming cooling optimization for
  green data center via deep reinforcement learning,'' \emph{IEEE Transactions
  on Cybernetics}, 2019.

\bibitem{tan2019cooperative}
T.~Tan, F.~Bao, Y.~Deng, A.~Jin, Q.~Dai, and J.~Wang, ``Cooperative deep
  reinforcement learning for large-scale traffic grid signal control,''
  \emph{IEEE Transactions on Cybernetics}, 2019.

\bibitem{Koch_2019}
\BIBentryALTinterwordspacing
W.~Koch, R.~Mancuso, R.~West, and A.~Bestavros, ``Reinforcement learning for
  uav attitude control,'' \emph{ACM Transactions on Cyber-Physical Systems},
  vol.~3, no.~2, pp. 1--21, Feb 2019. [Online]. Available:
  \url{http://dx.doi.org/10.1145/3301273}
\BIBentrySTDinterwordspacing

\bibitem{huang2017parameterized}
Z.~Huang, X.~Xu, H.~He, J.~Tan, and Z.~Sun, ``Parameterized batch reinforcement
  learning for longitudinal control of autonomous land vehicles,'' \emph{IEEE
  Transactions on Systems, Man, and Cybernetics: Systems}, vol.~49, no.~4, pp.
  730--741, 2017.

\bibitem{ding2018adaptive}
L.~Ding, S.~Li, H.~Gao, C.~Chen, and Z.~Deng, ``Adaptive partial reinforcement
  learning neural network-based tracking control for wheeled mobile robotic
  systems,'' \emph{IEEE Transactions on Systems, Man, and Cybernetics:
  Systems}, 2018.

\bibitem{cui2017adaptive}
R.~Cui, C.~Yang, Y.~Li, and S.~Sharma, ``Adaptive neural network control of
  auvs with control input nonlinearities using reinforcement learning,''
  \emph{IEEE Transactions on Systems, Man, and Cybernetics: Systems}, vol.~47,
  no.~6, pp. 1019--1029, 2017.

\bibitem{ferrari2004online}
S.~Ferrari and R.~F. Stengel, ``Online adaptive critic flight control,''
  \emph{Journal of Guidance, Control, and Dynamics}, vol.~27, no.~5, pp.
  777--786, 2004.

\bibitem{enns2003helicopter}
R.~Enns and J.~Si, ``Helicopter trimming and tracking control using direct
  neural dynamic programming,'' \emph{IEEE Transactions on Neural networks},
  vol.~14, no.~4, pp. 929--939, 2003.

\bibitem{zhou2018incremental}
Y.~Zhou, E.-J. van Kampen, and Q.~P. Chu, ``Incremental model based online dual
  heuristic programming for nonlinear adaptive control,'' \emph{Control
  Engineering Practice}, vol.~73, pp. 13--25, 2018.

\bibitem{wang2019deterministic}
Y.~Wang, J.~Sun, H.~He, and C.~Sun, ``Deterministic policy gradient with
  integral compensator for robust quadrotor control,'' \emph{IEEE Transactions
  on Systems, Man, and Cybernetics: Systems}, 2019.

\bibitem{xu2019morphing}
D.~Xu, Z.~Hui, Y.~Liu, and G.~Chen, ``Morphing control of a new bionic morphing
  uav with deep reinforcement learning,'' \emph{Aerospace Science and
  Technology}, 2019.

\bibitem{wu2018depth}
H.~Wu, S.~Song, K.~You, and C.~Wu, ``Depth control of model-free auvs via
  reinforcement learning,'' \emph{IEEE Transactions on Systems, Man, and
  Cybernetics: Systems}, vol.~49, no.~12, pp. 2499--2510, 2018.

\bibitem{shapiro2001using}
D.~Shapiro, P.~Langley, and R.~Shachter, ``Using background knowledge to speed
  reinforcement learning in physical agents,'' in \emph{Proceedings of the
  fifth international conference on Autonomous agents}, 2001, pp. 254--261.

\bibitem{Chen-2019-112809}
T.~Chen, ``Deep reinforcement learning with prior knowledge,'' Master's thesis,
  Pittsburgh, PA, May 2019.

\bibitem{gulccehre2016knowledge}
{\c{C}}.~G{\"u}l{\c{c}}ehre and Y.~Bengio, ``Knowledge matters: Importance of
  prior information for optimization,'' \emph{The Journal of Machine Learning
  Research}, vol.~17, no.~1, pp. 226--257, 2016.

\bibitem{ferranti2017value}
D.~Ferranti, D.~Krane, and D.~Craft, ``The value of prior knowledge in machine
  learning of complex network systems,'' \emph{Bioinformatics}, vol.~33,
  no.~22, pp. 3610--3618, 2017.

\bibitem{moreno2004using}
D.~L. Moreno, C.~V. Regueiro, R.~Iglesias, and S.~Barro, ``Using prior
  knowledge to improve reinforcement learning in mobile robotics,'' \emph{Proc.
  Towards Autonomous Robotics Systems. Univ. of Essex, UK}, 2004.

\bibitem{lillicrap2015continuous}
T.~P. Lillicrap, J.~J. Hunt, A.~Pritzel, N.~Heess, T.~Erez, Y.~Tassa,
  D.~Silver, and D.~Wierstra, ``Continuous control with deep reinforcement
  learning,'' \emph{arXiv preprint arXiv:1509.02971}, 2015.

\bibitem{yucelen2012low}
T.~Yucelen and W.~M. Haddad, ``Low-frequency learning and fast adaptation in
  model reference adaptive control,'' \emph{IEEE Transactions on Automatic
  Control}, vol.~58, no.~4, pp. 1080--1085, 2012.

\bibitem{gaudio2019connections}
J.~E. Gaudio, T.~E. Gibson, A.~M. Annaswamy, M.~A. Bolender, and E.~Lavretsky,
  ``Connections between adaptive control and optimization in machine
  learning,'' \emph{arXiv preprint arXiv:1904.05856}, 2019.

\bibitem{apkarian2006nonsmooth}
P.~Apkarian and D.~Noll, ``Nonsmooth {$H_\infty$} synthesis,'' \emph{IEEE
  Transactions on Automatic Control}, vol.~51, no.~1, pp. 71--86, 2006.

\bibitem{apkarian2015parametric}
P.~Apkarian, M.~N. Dao, and D.~Noll, ``Parametric robust structured control
  design,'' \emph{IEEE Transactions on Automatic Control}, vol.~60, no.~7, pp.
  1857--1869, 2015.

\bibitem{stein2016effect}
N.~Stein, H.~Weiss, G.~Hexner, and I.~Rusnak, ``Effect of missile configuration
  and inertial measurement unit location on autopilot response,'' \emph{Journal
  of Guidance, Control, and Dynamics}, pp. 2740--2745, 2016.

\bibitem{mracek2005missile}
C.~Mracek and D.~Ridgely, ``Missile longitudinal autopilots: comparison of
  multiple three loop topologies,'' in \emph{AIAA guidance, navigation, and
  control conference and exhibit}, 2005.

\bibitem{duan2016benchmarking}
Y.~Duan, X.~Chen, R.~Houthooft, J.~Schulman, and P.~Abbeel, ``Benchmarking deep
  reinforcement learning for continuous control,'' in \emph{International
  Conference on Machine Learning}, 2016, pp. 1329--1338.

\bibitem{henderson2018deep}
P.~Henderson, R.~Islam, P.~Bachman, J.~Pineau, D.~Precup, and D.~Meger, ``Deep
  reinforcement learning that matters,'' in \emph{Thirty-Second AAAI Conference
  on Artificial Intelligence}, 2018.

\bibitem{islam2017reproducibility}
R.~Islam, P.~Henderson, M.~Gomrokchi, and D.~Precup, ``Reproducibility of
  benchmarked deep reinforcement learning tasks for continuous control,''
  \emph{arXiv preprint arXiv:1708.04133}, 2017.

\end{thebibliography}

\begin{IEEEbiography}[{\includegraphics[width=1in,height=1.25in,clip,keepaspectratio]{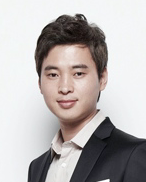}}]{Hyo-Sang Shin}
received his BSc on aerospace engineering from Pusan National University in 2004 and gained an MSc on flight dynamics, guidance and control in Aerospace Engineering from KAIST and a PhD on cooperative missile guidance from Cranfield University in 2006 and 2010, respectively. He is currently a Professor of Guidance, Control and Navigation Systems and Head of Autonomous and Intelligent Systems Group at Cranfield University. His current research interests include multiple target tracking, adaptive and sensor-based control, and distributed control of multiple agent systems.
\end{IEEEbiography}

\begin{IEEEbiography}[{\includegraphics[width=1in,height=1.25in,clip]{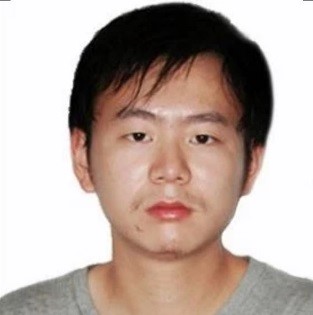}}]{Shaoming He}
received his BSc and MSc degrees in Aerospace Engineering from Beijing Institute of Technology, in 2013 and 2016, respectively, and a PhD in Aerospace Cranfield University. He is currently an associate professor in School of Aerospace Engineering at Beijing Institute of Technology. His research interests include multi-target tracking, UAV guidance and trajectory optimization.
\end{IEEEbiography}

\begin{IEEEbiography}[{\includegraphics[width=1in,height=1.25in,clip]{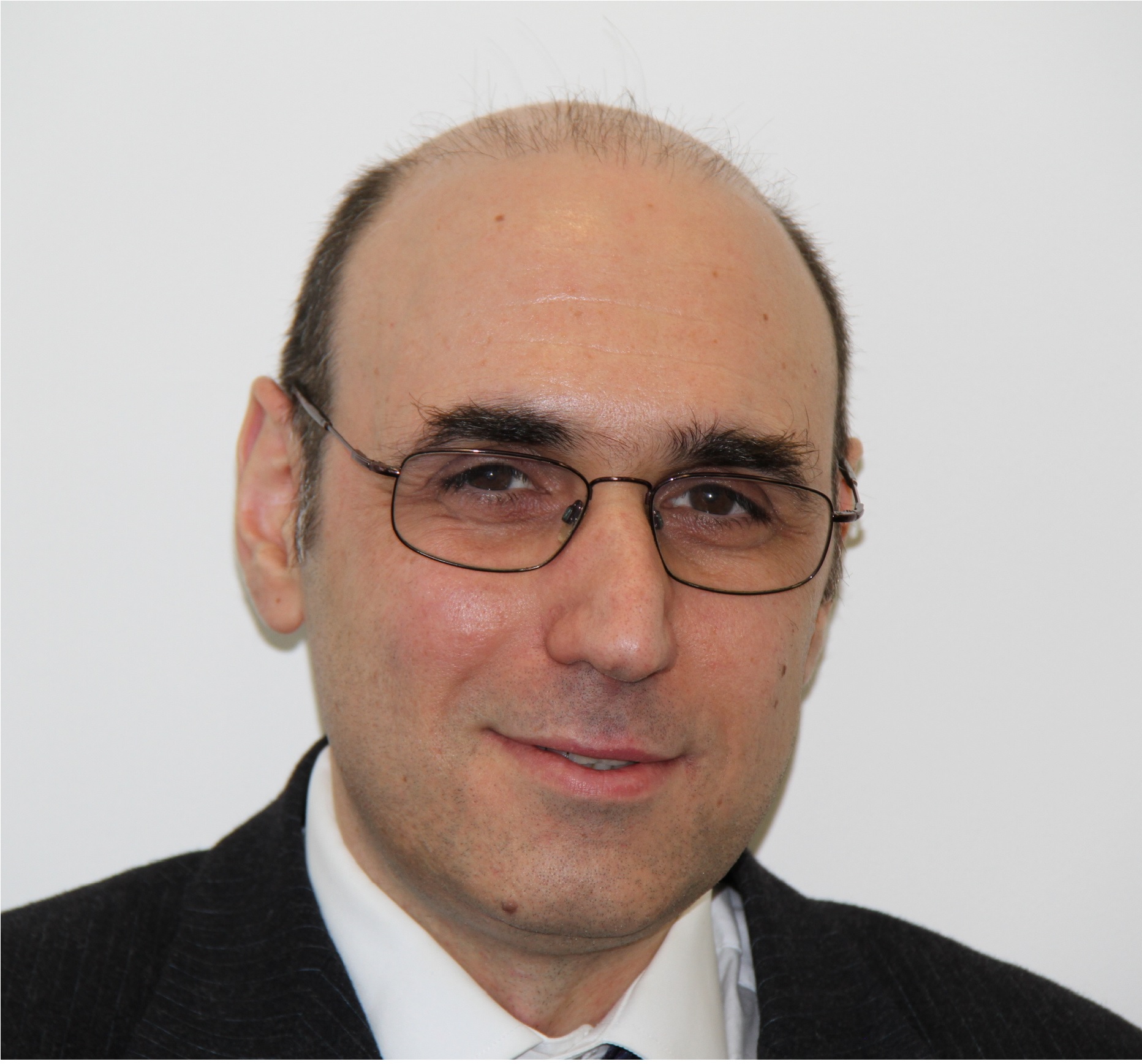}}]{Antonios Tsourdos} obtained a MEng in electronic, control and systems engineering from the University of Sheffield (1995), an MSc in systems engineering from Cardiff University (1996), and a PhD in nonlinear robust missile autopilot design and analysis from Cranfield University (1999). He is a Professor of Control Engineering with Cranfield University, and was appointed Head of the Centre for Cyber-Physical Systems in 2013. He was a member of the Team Stellar, the winning team for the UK MoD Grand Challenge (2008) and the IET Innovation Award (Category Team, 2009).
\end{IEEEbiography}

\end{document}